\documentclass{article}

\usepackage{arxiv}
\usepackage[utf8]{inputenc} % allow utf-8 input
\usepackage[T1]{fontenc}    % use 8-bit T1 fonts
\usepackage{hyperref}       % hyperlinks
\usepackage{url}            % simple URL typesetting
\usepackage{booktabs}       % professional-quality tables
\usepackage{amsfonts}       % blackboard math symbols
\usepackage{nicefrac}       % compact symbols for 1/2, etc.
\usepackage{microtype}      % microtypography
\usepackage{graphicx}
\usepackage{tikz}
\usetikzlibrary{shapes.geometric, arrows, calc, shadows}
\usepackage[disable]{todonotes}
\usepackage{caption}
\usepackage{textcomp}
\usepackage{gensymb}
\usepackage{siunitx}
\usepackage{amsmath}
\usepackage{listings}
\usepackage{subcaption}
\usepackage[titletoc]{appendix}
\usepackage[export]{adjustbox}
\newcommand{\clearemptydoublepage}{%
  \ifthenelse{\boolean{@twoside}}{\newpage{\pagestyle{empty}\cleardoublepage}}%
  {\clearpage}}
\title{Machine Learning-based optimal Mesh Generation in Computational Fluid Dynamics}

\author{
  Keefe Huang\\
  Technical University of Munich\\
  85748 Garching b. M\"unchen, Germany\\
  \texttt{keefehuang@gmail.com}\\
\And
  Moritz Krügener \\
  Technical University of Munich\\
  85748 Garching b. M\"unchen, Germany\\
  \texttt{moritz.kruegener@tum.de } \\
\And
  Alistair Brown\\
  Siemens Industry Software Computational Dynamics Limited\\
  W6 7NL London, UK\\
  \texttt{alistair.brown@siemens.com} \\
\And
  Friedrich Menhorn \\
  Technical University of Munich\\
  85748 Garching b. M\"unchen, Germany\\
  \texttt{menhorn@in.tum.de} \\
  \And
  Hans-Joachim Bungartz \\
  Technical University of Munich\\
  85748 Garching b. M\"unchen, Germany\\
  \texttt{bungartz@in.tum.de} \\
\And
  Dirk Hartmann\\
  Siemens AG, Technology\\
   81739 M\"unchen, Germany\\
  \texttt{hartmann.dirk@siemens.com} \\
}

\begin{document}
\maketitle

%%%%%%%%%%%%%%%%%%%%%%%%%%%%%%%%%%%%%%%%%%%%%%%%%%%%%%%%%%%%%%%%%
\begin{abstract}
Computational Fluid Dynamics (CFD) is a major sub-field of engineering. Corresponding flow simulations are typically characterized by heavy computational resource requirements. Often, very fine and complex meshes are required to resolve physical effects in an appropriate manner. Since all CFD algorithms scale at least linearly with the size of the underlying mesh discretization, finding an optimal mesh is key for computational efficiency. 

One methodology used to find optimal meshes is goal-oriented adaptive mesh refinement. However, this is typically computationally demanding and only available in a limited number of tools. Within this contribution, we adopt a machine learning approach to identify optimal mesh densities. We generate optimized meshes using classical methodologies and propose to train a convolutional network predicting optimal mesh densities given arbitrary geometries. The proposed concept is validated along 2d wind tunnel simulations with more than 60,000 simulations. Using a training set of 20,000 simulations we achieve accuracies of more than 98.7{\%}. 

Corresponding predictions of optimal meshes can be used as input for any mesh generation and CFD tool. Thus without complex computations, any CFD engineer can start his predictions from a high quality mesh. 
\end{abstract}


%\newpage
%\tableofcontents
%\newpage
%%%%%%%%%%%%%%%%%%%%%%%%%%%%%%%%%%%%%%%%%%%%%%%%%%%%%%%%%%%%%%%%%
\section{Introduction}\label{sec:introduction}
Computer Aided Engineering (CAE) tools are integral parts of industrial product development today. Among these, Computational Fluid Dynamics (CFD), addressing the simulation and prediction of fluid and air flows as well as their effects, is the fastest growing domain. One of the major limitations of CFD is its demand in terms of computational resources. The accuracy of the prediction depends very much on the specific mesh, e.g. a very fine mesh in regions where boundary layer separation sets in or turbulent structures develop. Identifying an optimal mesh is a highly complex task and particularly depends on the specific flow situations. Years of experience are typically required to be able to generate efficient computational meshes.

Several strategies have been proposed in the past to determine optimal computational meshes \cite{plewa2005adaptive}. Among these, refinements based on goal-oriented error estimators \cite{becker2001optimal} have been the most successful ones. Given a simulation goal, e.g. drag calculation, they determine which regions are likely to contribute the biggest error and refine the mesh in these regions correspondingly. Performing this task iteratively Pareto optimal meshes in terms of accuracy and degrees of freedom can be found. However, this calculation itself is highly computationally demanding and only available in a limited number of computational tools. 

In the past years, a number of machine learning technologies have been proposed to accelerate the process of flow simulations, by either learning from corresponding simulations and predict the results by means of neural networks \cite{Guo2016} or by combining both approaches and thus accelerating simulations \cite{Ling2016}. However in the latter case, the identification of optimal computational meshes is not resolved. Furthermore, a major concern of these approaches in the context of industrial use is that any machine-learning based approach will be hard to be validated. 

Within this contribution, we adopt a different approach. Instead of accelerating the flow simulation itself we focus on a machine-learning-based solution to accelerate optimal grid generation. Following the approach of \cite{Kutz2017}, we determine optimal meshes for a large number of specific situations using a commercial solver (Simcenter STAR-CCM+ \cite{StarCCM}). Based on a large set of examples / data points we train a convolutional neural network to predict optimal mesh distributions. The predicted optimal mesh densities can be used as an input for any CFD meshing/ simulation tool. Here, we focus on the task of identifying an optimal mesh for a 2-dimensional channel/ wind-tunnel-like flow with arbitrary geometries inside.  A generalization to other examples is straight forward.

Such kind of mesh refinement based on machine learning is a field with high likelihood for industrial adoption. First, compared to many previous works combining machine learning and simulations, the approach is not sensitive to validation and verification. Second, while non-optimal meshes can lead to bad convergence behavior, they still result in correct predictions. Third and finally, only a limited set of industrial simulation tools provide solutions for the adjoint problem while a lot of adaptive mesh refinement strategies rely on them. The predicted mesh---trained using adjoint solutions---is computationally inexpensive to create and can directly be used or provide an initial starting guess for further refinement.

In the following sections we present our new approach for machine-learning based optimal mesh generation. In \autoref{sec:SOA} we discuss the current state of the art and give a recap on adaptive mesh refinement as well as a brief history on how machine learning is used in CFD. In \autoref{sec:DataPipeline} we present how we create the data which is later on used for training the convolutional neural network which is introduced in \autoref{sec:TrainingPipeline}. Afterwards we show in \autoref{sec:Results} results for the training as well as prediction of two sample geometries---one from the test set and one randomly generated which has not been seen during training or validation. We close this paper with a discussion in \autoref{sec:Conclusions} on how this approach can be used in CFD.

%%%%%%%%%%%%%%%%%%%%%%%%%%%%%%%%%%%%%%%%%%%%%%%%%%%%%%%%%%%%%%%%%
\section{State of the Art}
\label{sec:SOA}
Current CFD methodologies allow for very accurate results to be produced at the price of high computational cost. This can lead to simulation times lasting days or even weeks for large and complex geometries. As a result, design optimization studies are often too costly and slow to be feasible.

Simulation time can be reduced by accepting a higher error or coarser resolutions but this can lead to non-physical behaviour in the simulation. The goal of introducing adaptive mesh refinement and a corresponding machine learning version is to achieve a speed-up of the overall process without sacrificing in terms of accuracy and resolution. Therefore, in the next section we will give a review on adaptive mesh literature and approaches for machine learning in CFD.

% % % % % % % % % % % % % % % % % % % % % % % % % % % % % % % % %
\subsection{Adaptive Grid Refinement}
\label{sec:SOA_AGR}
Since the first days of continuum mechanics simulations, adaptive mesh refinement has been in the focus of research. The computation time of corresponding simulations scales at least linearly with the size of the mesh, i.e. the number of vertices or degrees of freedom, but in most cases even quadratically. Since in many situations the physical quantity of interest is heavily influenced by local phenomena, having an effective non-homogeneous adaptive grid with an appropriate local refinement, i.e. an adaptivity in mesh size, is key for economic simulations. Choosing an appropriate computational mesh very often depends on the experience of the simulation expert using corresponding tools. Though over the years quite a number of heuristic approaches have been developed \cite{plewa2005adaptive}.

One approach to refine the mesh is to use classical error estimation methods, e.g. estimating errors in global (energy) norms exploiting variational formulations of the underlying problems \cite{babuvska1987feedback}. However, these kind of global bounds are often agnostic of the actual physical quantities of interest, e.g. drag of an object or a physical value at a specific location. That is, bounds on the errors of the quantities of interest cannot be related directly to the global estimates. This makes it difficult to get indications where and how to refine the computational mesh.

Dual Weighted Residual (DWR) error estimators take a different approach \cite{becker2001optimal}: They estimate local residuals of the numerical solutions and estimate their effect on the physical quantities of interests. The latter is achieved by solving an adjoint problem identifying sensitivities of the quantities of interest on the local error, i.e. they allow to estimate and localize the error.  Thereby, the functional of interest cannot only be norms but also highly specific values such as any integral or point quantities. Realizing a feedback loop, the computational grid can be iteratively refined achieving an optimal mesh.  This approach does not only allow for an efficient a posteriori error control for the physical quantities of interests but at the same time also results in highly economical meshes with optimal efficiency for the functional of interest.

The DWR method goes back to Becker and Rannacher \cite{becker1996weighted} and is based on the pioneering work of Babu{\v{s}}ka and Rheinboldt \cite{babuvvska1978error, babuvska1978posteriori} and later refined by many others as Eriksson, Estep, Hansbo and Johnson  \cite{eriksson1995introduction}. For a complete overview we refer e.g. to the following surveys \cite{verfurth1994posteriori, ainsworth1993unified}. Though, originally introduced for Finite Element Methods the DWR is versatile can can be generalized to other discretizations as well, e.g. Finite Volume Methods \cite{barth2005posteriori}. Corresponding functionality is available in academic as well as industrial simulation tools, e.g. deal.II \cite{dealII}, FEniCS \cite{FEniCS}, Simcenter STAR-CCM+ \cite{StarCCM}, or Ingrid Cloud \cite{IngridCloud}, and has been applied for a large number of different applications, including fluid-dynamics \cite{becker2002optimal}, structural dynamics, as well as to complex multi-physics problems like chemically reactive flows \cite{braack2004adaptive} or fluid–structure interactions \cite{gratsch2006goal}.

The usage of these methods, however, is rather limited in industrial practice today despite the opportunity of allowing for optimal meshes, i.e. highly efficient computations, and reliable error estimates. They are inherently based on the solution of adjoint problems which requires dedicated functionality of corresponding simulation tools. In particular, only a limited set of industrial simulation tools provides a corresponding capability. To circumvent that, we propose to learn optimal meshes trained on corresponding solvers with adjoint functionality which can be used as an inital starting point also in other simulation tools independent of the specific numerical method. In the next section, we review how machine learning techniques are applied in CFD as of now.

% % % % % % % % % % % % % % % % % % % % % % % % % % % % % % % % %
\subsection{Machine Learning in CFD}
\label{sec:SOA:ML_CFD}
Many methodologies exist for CFD. Similarly, machine learning techniques can be incorporated in a multitude of ways. The most straight-forward method is to replace the CFD simulation step altogether: training a network using the geometry and flow parameters in order for it to output the resulting flow.

The results of these replacement networks are highly dependent on the amount and quality of data used for training as well as the flow setup it is trained for. For example, Guo et. al. \cite{Guo2016} were quite successful in deploying this technique for steady state laminar flows. They report a speedup of at least 2 orders of magnitude while keeping the overall error increase moderate for most scenarios. However, this result does not hold for all flow setups.

Employing machine learning techniques for turbulent flows, however, has proved challenging, e.g., the straight forward idea to deploy one network to replace an entire turbulent flow solver was not successful \cite{ thuerey2018, brunton2020machine}. There are, nevertheless, approaches that yield approximate results with higher throughput by replacing the most expensive solver step, the pressure step, with a neural network \cite{tompson2017accelerating}. Another example is Ling et. al. \cite{Ling2016} who used a modified a RANS solver to include a network predicting the Reynolds stress anisotropy tensor with encouraging results. Furthermore, Kutz \cite{Kutz2017} suggests that especially deep neural networks can yield high accuracy and throughput results as they are better suited for the high-dimensional input data and could potentially capture very complex flow phenomena in the deeper layers of the network.

These are only a few examples in a rapidly growing field. For a more extensive summary we refer to the recent review \cite{brunton2020machine}.

%%%%%%%%%%%%%%%%%%%%%%%%%%%%%%%%%%%%%%%%%%%%%%%%%%%%%%%%%%%%%%%%%
\section{Data Generation Pipeline}
\label{sec:DataPipeline}
One of the most important requirements for proper training is a representative set of training data. In this work, we employ supervised learning. Hence, we need to generate input-output pairs with ground truth results. This is accomplished via a data generation pipeline consisting of several key parts: First, we generate random geometries, described in  \autoref{sec:DataPipeline:GeometryGeneration}. Then, an iterative refinement process is employed using repeated primal (\autoref{sec:DataPipeline:PrimalSolve}) and adjoint (\autoref{sec:DataPipeline:DualSolve}) solves. The results of the adjoint solves are used to refine the mesh at the end of each iteration, using the process described in  \autoref{sec:DataPipeline:Adaptivity}. The data generated by the pipeline can then be processed for training.

% % % % % % % % % % % % % % % % % % % % % % % % % % % % % % % % %
\subsection{Random Geometry Generation}
\label{sec:DataPipeline:GeometryGeneration}
\begin{figure}
  \centering
  \includegraphics[width=0.95\linewidth]{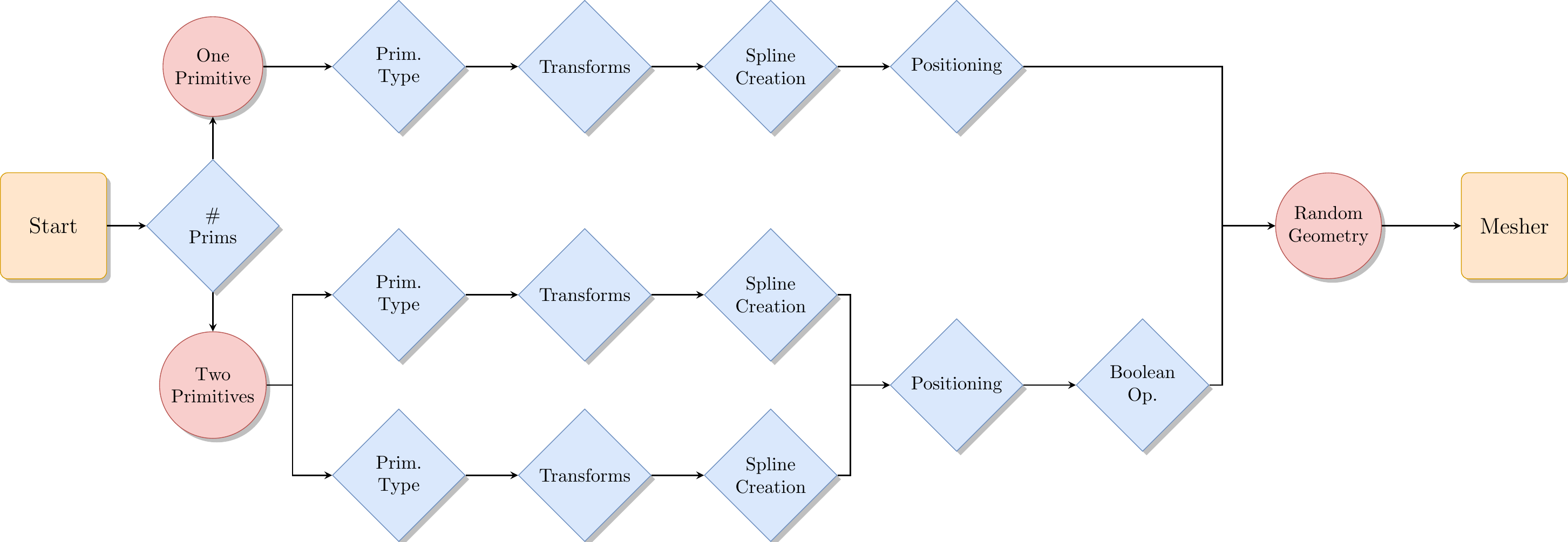}
  \captionof{figure}{Geometry creation pipeline}
  \label{fig:geometry_creation_pipeline}
\end{figure}

The aim of geometry generation is to generate a randomized set of geometries that is representative of the full space of possible CFD simulations within the selected constraints. Corresponding simulations will be the basis for generating data-sets to be used for the machine learning part. We divide the general geometry creation process into several \textit{phases}---each with a randomized component. These phases are illustrated as diamonds in the flowchart in figure \ref{fig:geometry_creation_pipeline}. Additionally, we show an example for the occurring transformations in the column on the right starting from two primitives---a triangle and a square. Note that in this work, we restrict the geometry to a $L \cdot L$ square region within the larger flow domain. 

The first phase in geometry creation is the \textit{primitives phase}. In this phase, the number of primitives is randomly determined. We use five basic primitives: a triangle, square, pentagon, hexagon, and duodecagon. This extends the space of possible created geometries since relying on a single geometry cannot produce representative results. Currently, the pipeline only supports the use of one or two primitives. 

The second phase is the \textit{primitive type selection}. Our approach mirrors the methodology applied in \cite{Ling2016}, where several primitives for random geometry creation were used when training a neural network on a similar problem.  Unlike in \cite{Ling2016}, we omit the circle or ellipse.  Instead we smooth some of the sharp corners in primitives by replacing lines between vertices by splines. We employ a uniform distribution, i.e. there is an equal chance that any one geometry is selected.

The third phase is the \textit{transform phase}. In this phase, several transformations are performed on base primitives to create a randomized shape. We begin with a uniform, unit primitive, such as a unit square. Then, two extensions, one rotation and a displacement are performed. An extension has two random variables, the axis of extension and the extension factor. The axis of extension is an axis with an angle $\alpha$ from the horizontal plane between 0 and $180\degree$ that passes through the centroid of the base primitive. Each vertex is then extended by the extension factor $\lambda$ in the direction normal to the axis of extension $\vec{n}$.  While the axis of extension has a uniform distribution between 0 and 180\degree, the extension factor is normally distributed between 1 and 5, with a mean of 3 and a sigma of 1.5. 

In the fourth phase, the \textit{positioning phase}, the geometry is randomly rotated and then positioned within the specified bounds. The first primitive is positioned in the middle of the allowed bounds to avoid any shift of the flow towards boundaries. The second primitive (if one is used) is positioned randomly until both overlap. Thereby a uniform distribution is used to determine the random position of the second primitive. 

The fifth phase is the \textit{spline creation phase}. The use of splines is suggested in the place of straight lines between vertices to reduce the number of sharp edges. We implemented this by placing midpoints randomly between vertices and drawing splines using the new set of points.  There are two random aspects when performing this operation: the number of midpoints and their random distribution between the two vertices. Here, we use 0-5 midpoints with a uniform  distribution.
	
The final phase in random geometry creation is to \textit{merge  transformed primitives} by performing a boolean operation. This phase only takes place if two primitives are used. There are two possible boolean operations---union and intersection. A union combines the two primitives while an intersect uses only the region where the two primitives intersect. Note that the intersect operation introduces some complexity. Firstly, there is a general tendency to form slivers. Secondly, there is also no current control on the minimum size of intersected region. To prevent arbitrary small sizes of intersecting regions, we extend all intersect regions by a random factor uniformly chosen between 3-5 after the intersect operation is performed.

% % % % % % % % % % % % % % % % % % % % % % % % % % % % % % % % %
\subsection{CFD Simulation: Primal Solve}
\label{sec:DataPipeline:PrimalSolve}
After having a pipeline to generate random geometries, the next step is to set-up the primal CFD simulation to allow for convergence with a wide variety of random geometries. This section overviews the inputs for the primal solver, the adjoint solver and the resulting mesh refinement. We employ  Simcenter STAR-CCM+ \cite{StarCCM}, a commercial CFD software developed by Siemens Digital Industry Software, for the simulations.

% % % % % % % % % % % % % % % % % % % % % % % % % % % % % % % % %

\subsubsection{Computational Domain}
Since turbulence effects extend extremely far from the base geometry, we place the generated geometry in a region of size $L \cdot L$ which is positioned $75L$ away from the inlet as well as the upper and lower boundaries and $225L$ away from the outlet as illustrated in figure \ref{fig:cfd_mesh}. The resultant domain size has a blockage ratio of less than $0.0005\%$ to ensure the free slip boundary conditions. For this experiment, we set the length $L$ to be $5m$, with an appropriately scaled domain. This domain is fully within the turbulent regime, such that it serves as a starting point for similar projects involving fast moving medium-sized objects, like moving vehicles or wind-tunnel-like applications.

The boundary conditions applied to the flow problem are depicted on the left side of figure \ref{fig:cfd_mesh}.
\begin{enumerate}
    \item At the \textit{inlet boundary}, the flow is defined as subsonic and incompressible with an inflow velocity
    perpendicular to the boundary wall. An inflow velocity of $17.6 m/s$ was used, which is approximately $60 km/h$.
    \item At the \textit{outlet boundary}, the flow regime is also defined as subsonic, with a static pressure of $1 \si{atm}$.
    \item The \textit{upper} and \textit{lower wall boundaries} are considered as far-field and defined as free-slip walls.
    \item The randomly generated geometry or \textit{obstacle} is defined as a no-slip wall with the wall roughness set to smooth.
\end{enumerate}

% % % % % % % % % % % % % % % % % % % % % % % % % % % % % % % % %

\subsubsection{Initial Mesh Setup}
Though the goal of this work is to learn the mesh sensitivity with respect to certain input geometries, it is still necessary to have a mesh with good quality to improve the robustness of the mesh refinement process.

\begin{figure}[htp!]
  \centering
  \includegraphics[width=0.69\textwidth,valign=c]{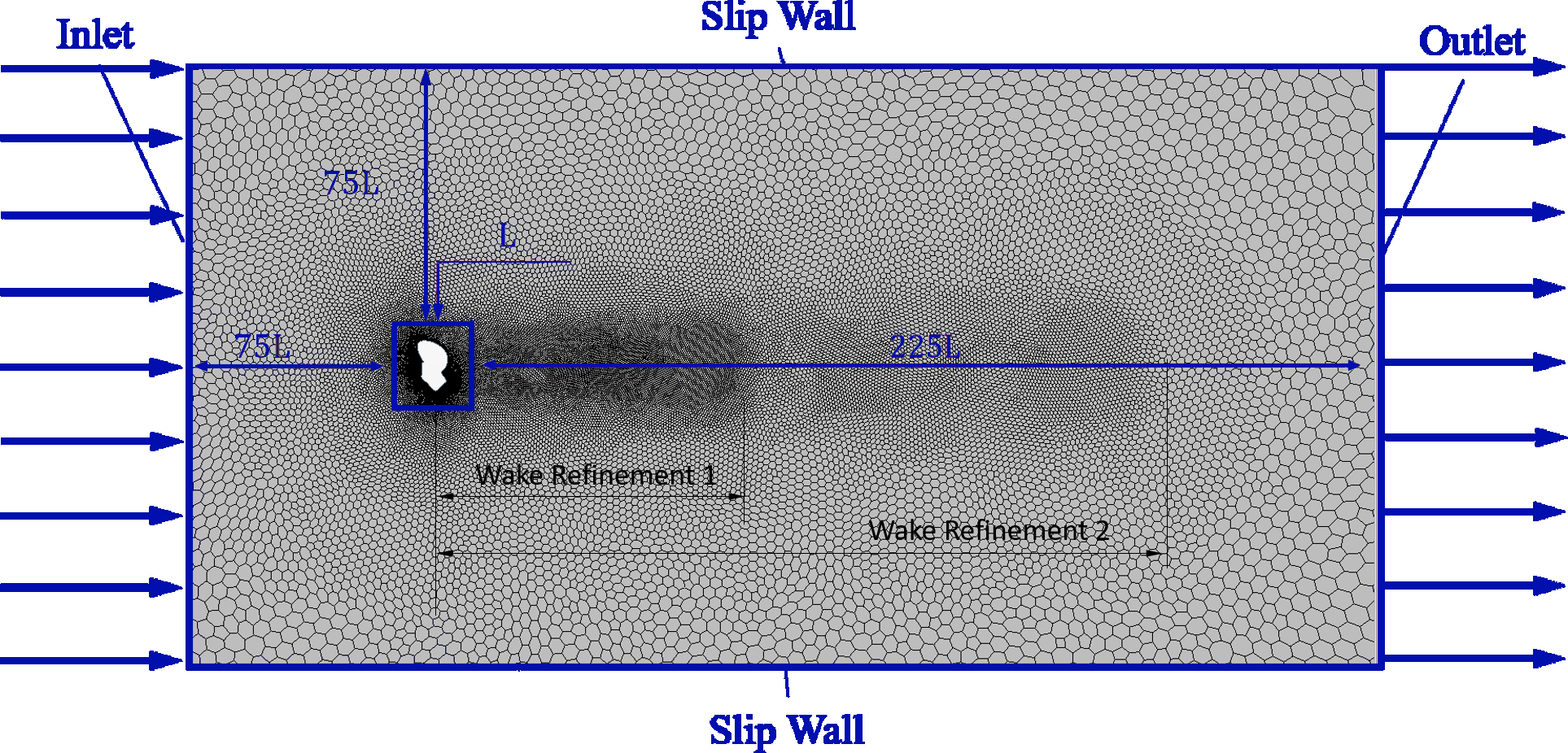}\hfill
  \includegraphics[width=0.24\textwidth,valign=c]{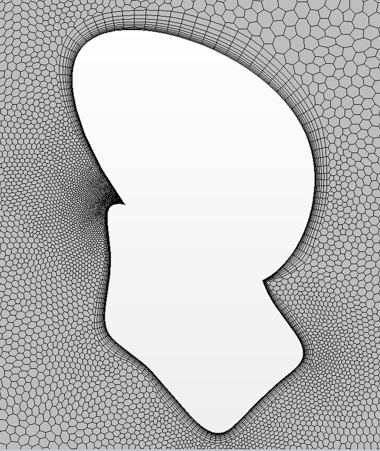}
  \caption{Global mesh overview of a random geometry: complete setup and close-up of the geometry and grid. (Note that the illustration is not in scale to also illustrate the mesh refinement.)}
  \label{fig:cfd_mesh}
\end{figure}

A meshing example is illustrated in figure \ref{fig:cfd_mesh}. We employ polygonal mesh elements as the Simcenter STAR-CCM+ solver is optimized for this element type. Additionally, two wake refinements are defined to capture the eddies behind the obstacle. Prism layers are employed at the obstacle boundary. Detailed parameters for the setup of the mesh including prism layers and wake refinement can be found in appendix \ref{app:CFD_solver_mesh}.

% % % % % % % % % % % % % % % % % % % % % % % % % % % % % % % % %

\subsubsection{Primal Solver}
To make our CFD runs as fast but also as robust as possible we set the following solver specifications. We employ a coupled implicit scheme due to its robustness. Furthermore, the Algebraic Multigrid (AMG) method is used to accelerate the convergence. To speed up the convergence further, we use a large CFL number of 200. To avoid instability in the initial iteration steps, we employ linear CFL ramping for the first 50 solver iterations, starting with a CFL number of 0.25. Additionally, we use grid sequencing to provide a better "guess" of the flow solution. The detailed parameters for grid sequencing initialization can be found in appendix \ref{app:CFD_solver_primal}.

To ensure an efficient use of the computing ressources, we furthermore define several stopping criteria to allow the easy automation of running CFD simulations. The goal of these criteria is to stop simulations once we reach the desired accuracy or if the simulation does not converge. There are three situations where the simulation should terminate: reaching steady state, reaching sufficient accuracy, or  reaching maximum iteration number. The three cases as well as the associated parameters are detailed in appendix \ref{app:CFD_solver_primal}.

% % % % % % % % % % % % % % % % % % % % % % % % % % % % % % % % %
\subsection{CFD Simulation: Adjoint Solve}
\label{sec:DataPipeline:DualSolve}
Once the primal solution converges, the adjoint solver is employed. In general, the adjoint solution is an efficient way to compute gradients of a quantity of interest based on the input parameters. For a brief introduction we refer to the appendix \ref{app:adjoint} and to \cite{Giles2000}. In our specific case, we can use the resulting gradients to compute sensitivites of the input quantities with respect to cell densities of the current mesh. This helps us to identify areas where mesh refinement is needed to reduce the numerical error of the computed drag force. We refer to the calculated sensitivity as the estimated adjoint error or \textit{mesh sensitivity}.

First, the cell residuals of the Navier-Stokes equations are recomputed with a higher-order discretization scheme. The difference between the primal solution and the higher order residual is taken as an estimation of the true numerical error. Next, the field of residual differences is multiplied with the adjoint solution of the given cost function to obtain the mesh sensitivity. Thereby, adjoint methods significantly reduce the cost of calculating parametric sensitivities by selecting a specific Lagrange multiplier. For a short explanation we refer to appendix \ref{app:adjoint}. The main savings of the adjoint method stem from the lower complexity of solving the adjoint equation. Let us assume that solving for $R$ has a complexity of $M$, and solving the adjoint equations has a complexity of $N$. If $M \gg N$, when investigating a set of x parameters, implementing the adjoint equations reduces the complexity of sensitivity calculation from $O\left( \left(x + 1\right) M \right)$ to $O\left(M + xN\right) \approx O\left(M\right)$ \cite{Giannakoglou2008}.

\begin{figure}[ht]
  \centering
  \includegraphics*[width=10cm]{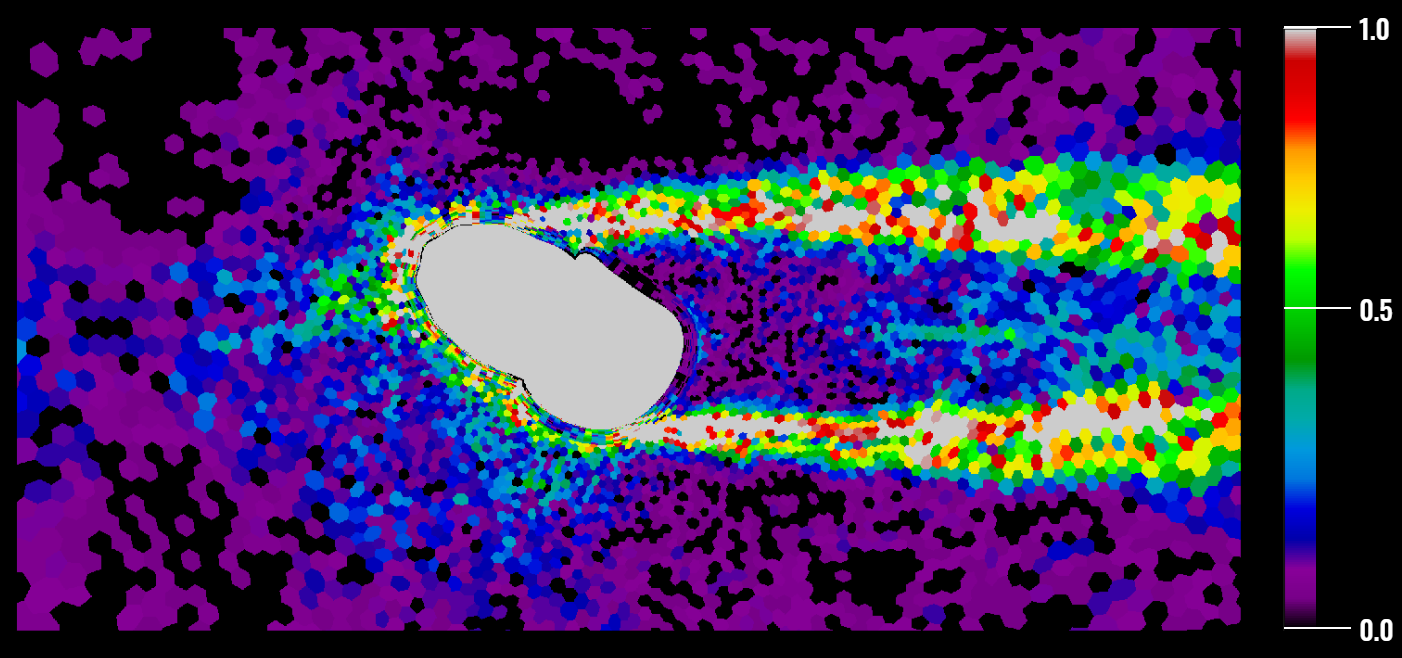}
  \caption{Mesh sensitivity estimation}
  \label{fig:adjoint_error}
\end{figure}

In this contribution, the drag force on the obstacle surface is chosen as the cost function. Since the magnitude of the error varies from geometry to geometry, we normalize the adjoint error by the drag force. An example of mesh sensitivity is shown in figure \ref{fig:adjoint_error}. 

% % % % % % % % % % % % % % % % % % % % % % % % % % % % % % % % %
\subsection{CFD Simulations with Adaptively Refined Meshes}\label{sec:DataPipeline:Adaptivity}
After obtaining the adjoint error, the estimated error per cell $\epsilon(i)$ is checked and compared to a threshold value, e.g. $\epsilon(i) > 0.1 \max(\text{adjoint error estimate})$. If the error is above this value, we refine the affected cell by reducing the base size of the cells in the local region. We note that when a cell is refined, the neighbouring cells are also adapted to generate a continuous mesh based on the growth ratio. This ratio determines that maximum change in neighboring cell sizes. A growth ratio of $1.17$ was chosen to ensure smooth transitions in the domain after refinement.

\begin{figure}[htp!]
  \centering
  \includegraphics*[width=10cm]{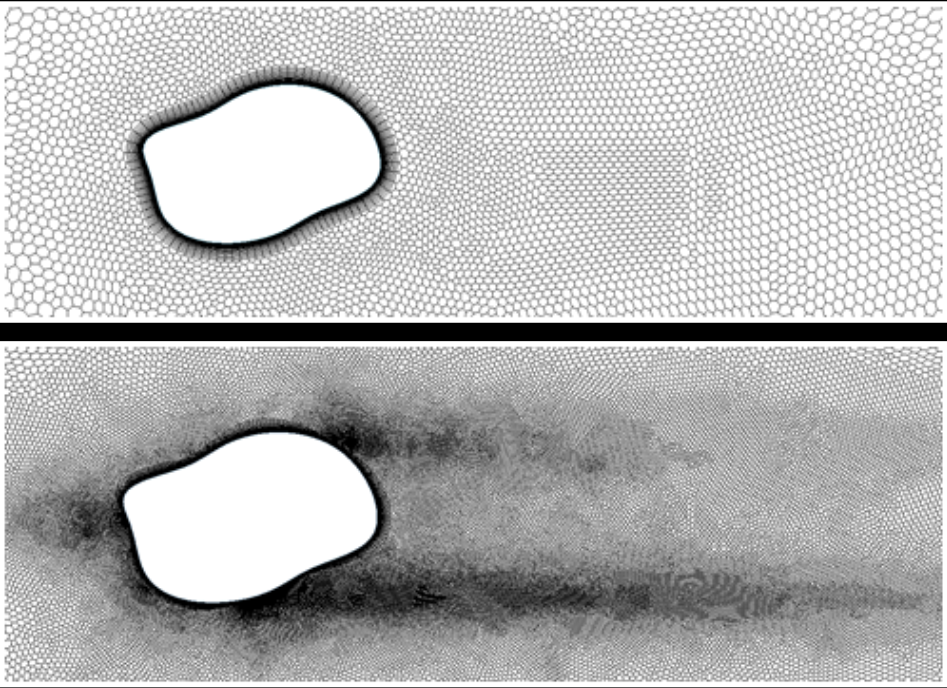}
  \caption{Top: No mesh refinement. Bottom: After several refinement steps}
  \label{fig:mesh_refinement}
\end{figure}

After refining the mesh, we then obtain a new adjoint solution by repeating the primal and adjoint solve using the newly refined mesh. For the purpose of data generation, we repeat this refinement process until either the adjoint solution reaches a steady state or the adjoint solver no longer converges. This means that both a new primal solve and adjoint solve is required for each refinement step. An example of a mesh after multiple refinements can be seen in figure \ref{fig:mesh_refinement}.

%%%%%%%%%%%%%%%%%%%%%%%%%%%%%%%%%%%%%%%%%%%%%%%%%%%%%%%%%%%%%%%%%
\section{Neural Network Training Pipeline}\label{sec:TrainingPipeline}
CFD simulations are usually based on non-rectilinear mesh structures with varying cell sizes in the simulation domain. The spatial irregularity of the mesh poses a difficult prediction target. As the number of cells and their size vary between simulations, predicting individual edge or node positions via a neural network approach is infeasible. Fortunately, the user is typically not interested in the exact position of a specific edge in the mesh, but rather in a typical cell size in a given area of the simulation domain. We therefore chose the average size of cells in a region as the prediction target---essentially corresponding to the refinement level of the mesh. This quantity can easily be described in a rectilinear grid covering the domain of interest and is not concerned with individual cells, which there can be hundreds of thousands of. The regular nature of this grid also lends itself to techniques used in image processing. We therefore chose to export the cell sizes from our CFD solver as an image, with the gray scale channel indicating the relative size of cells. When these images are blurred and down sampled as described in \autoref{sec:DataPipeline:DataPreparation}, the resulting 128x128 images are a representation of mesh refinement and are easily fed into a convolutional neural network as prediction targets.

In the following, we detail the overall processes and architectures chosen in this work. The architecture of choice, the \textit{Staircase UNet Architecture}, is introduced in \autoref{sec:TrainingPipeline:Architecture}. The pre-processing of data for training is explored in \autoref{sec:DataPipeline:DataPreparation}.

% % % % % % % % % % % % % % % % % % % % % % % % % % % % % % % % %
\subsection{Neural Network Architecture}\label{sec:TrainingPipeline:Architecture}
\begin{figure}[htp]
  \centering
  \includegraphics[width=0.35\linewidth]{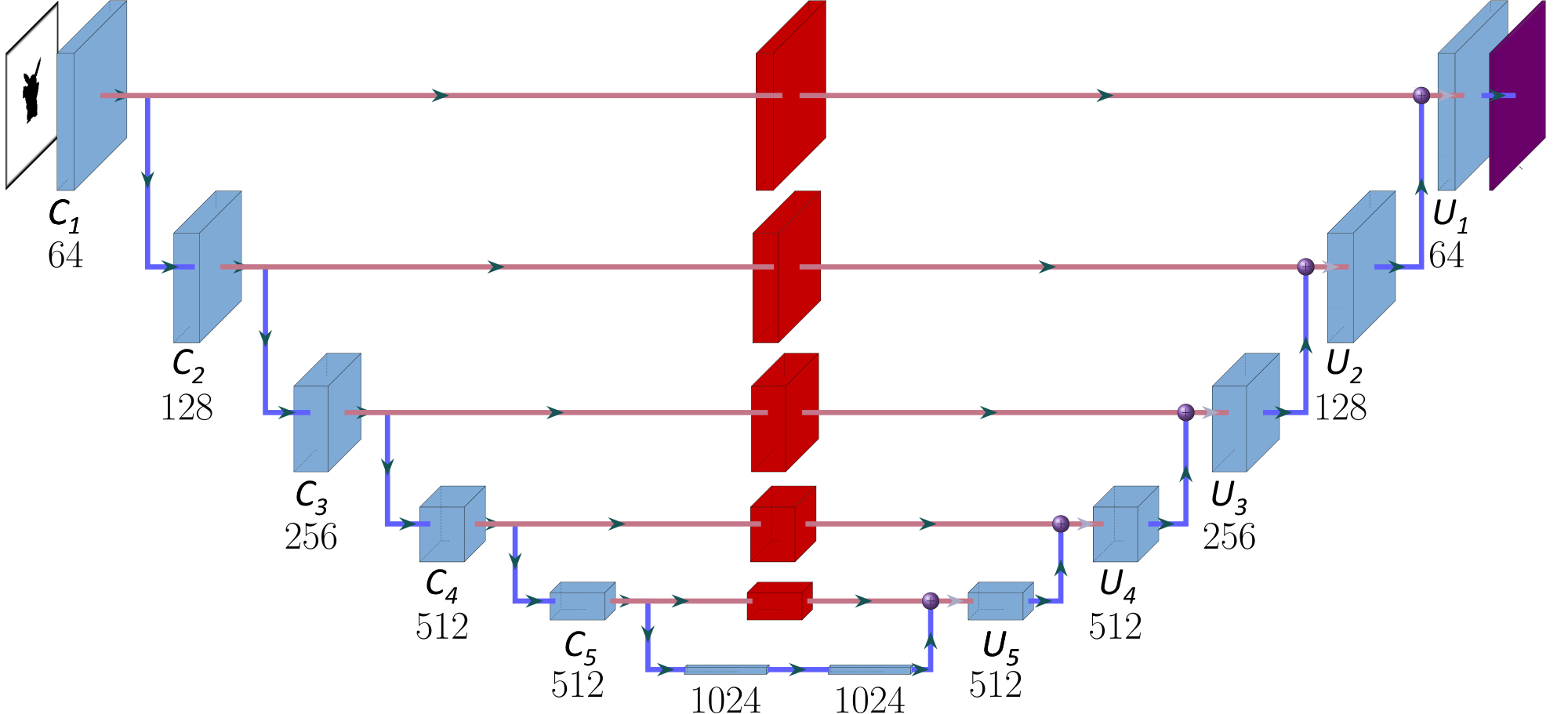}
  \hspace{0.05\linewidth}
  \includegraphics[width=0.5\linewidth]{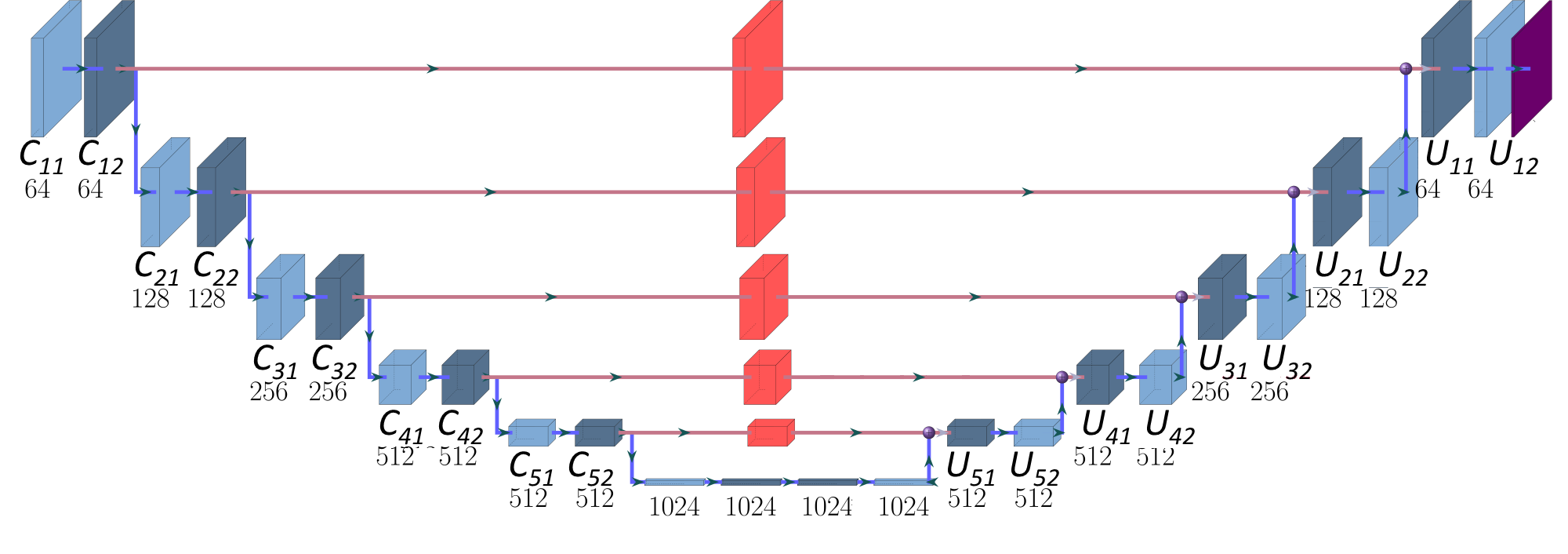}
  \caption{Left: UNet architecture (blue) with skip-connections (red); Right: Staircase UNet architecture with skip-connections}
  \label{fig:unetskip}
\end{figure}

The UNet architecture with skip-connections used in \cite{thuerey2018} is also used in this work. This is a simple convolutional neural network consisting of multiple layers. In each layer, input data passes through a convolution, and an activation function followed by down- or up-sampling. Additional steps can be taken for improved performance, such as batch normalisation or dropout but these  will be discussed later when addressing hyper-parameter tuning. A basic UNet architecture without skip-connections is shown in figure \ref{fig:unetskip} in blue. Here, the number of convolutional channels in each layer increases as the data is down-sampled and decreased as it is up-sampled.  As the size of our input and output data is the same, we consider only symmetric UNet architectures.

In addition to the basic UNet structure, the authors of \cite{thuerey2018} also utilized skip-connections  and successfully predicted flow-based quantities. These connections can be seen in figure \ref{fig:unetskip}, represented by light red lines.
 
Skip-connections allow for information to be transferred between certain input and output layers in a neural network \cite{mao2016}. They are important because the magnitude of the coefficients in the initial layers of a neural network are often much smaller compared those in the final layers. As a result, the gradients of the loss in these initial layers is often very small, which typically results in a poorly-conditioned optimization problem with slow convergence. Skip-connections alleviate this problem by equalizing the magnitudes of the gradients since they reduce the effective distance between output and input layers, thereby improving the training  of much deeper networks \cite{mao2016}.

In this work, we modify the UNet with skip-connections model, creating a \textit{Staircase UNet Architecture}. This structure adds additional layers at each depth of the network, such that additional convolutions are performed before up- or down-sampling occurs, a visual representation of this architecture can be seen in figure \ref{fig:unetskip}. 

With this architecture, we aim to preserve more information when passing data through skip-connections. This is best understood by examining figure \ref{fig:unetskip}. When utilizing the staircase architecture, connecting layers $C_{21}$ and $U_{22}$ would be analogous to utilizing the basic skip connection seen in figure \ref{fig:unetskip} because the combined data from these two layers would be immediately upsampled. However, if we chose to connect layers $C_{21}$ and $U_{21}$ or $C_{22}$ and $C_{21}$, the resulting combined data would undergo another full convolutional layer before upsampling---the idea being that upsampling could destroy features that would otherwise prove useful for training. Henceforth, when referring to these two types of skip-connections, we call the former, where information is immediately upsampled, an \textit{outer skip-connection} and the latter, which allows for an additional convolution prior to upsampling, an \textit{inner skip-connection}.

% % % % % % % % % % % % % % % % % % % % % % % % % % % % % % % % %
\subsection{Data Preparation}\label{sec:TrainingPipeline:Data}
\label{sec:DataPipeline:DataPreparation}
\subsubsection{Smoothing}
Our input images often contain sharp edges where the values of adjacent cells differ greatly. Additionally, capturing individual cells, which in practical CFD applications can have arbitrary shapes and sizes, is difficult in machine learning. Hence, a Gaussian blur (commonly used in graphics and computer vision) is used to smoothen the edges and improve training. That is, we apply a convolution to the image with a Gaussian function:
\begin{equation}
  \label{eq:Gaussian}
   G_{\sigma}\left(x, y\right) = \frac{1}{2  \pi  \sigma^2} exp\left(-\frac{x^2 + y^2}{2 \sigma^2}\right),
\end{equation}
which depends on parameters $\sigma$ and radius $r$. Thus for a pixel at $\left(x, y\right)$ the Gaussian blur $B_{\sigma, r}\left(x, y\right)$ of a pixel $(x,y)$ is defined as follows:
\begin{align}
  \label{eq:GaussianBlur}
  B_{\sigma, r}\left(x, y\right) &= \frac{1}{k}\sum_{rx=-r}^{r}\sum_{ry=-r}^{r} G_{\sigma}\left(rx, ry\right) F\left(x + rx, y + ry\right)\\
  \label{eq:GaussianBlurK}
  \text{with} \enspace k &= \sum_{rx=-r}^{r}\sum_{ry=-r}^{r} G_{\sigma}\left(rx, ry\right),  
\end{align}
where $F\left(x, y\right)$ is a function of pixel value in the original image. When performing the blur, we apply Neumann boundary conditions and assume that the domain beyond the picture is of the same color as that on the boundary.

Additionally, we cannot apply this blur onto geometry pixels as this would destroy geometry boundary information. Hence, we apply a selective blur that ignores data on the geometry. This is done by modifying the convolution kernel for each pixel such that it excludes data, thus restricting smoothing to non-geometry pixels:
\begin{align}
  \label{eq:FinalKernel}
  B_{\sigma, r}\left(x, y\right) &= \frac{1}{k\left(x, y\right)}\sum_{rx=-r}^{r}\sum_{ry=-r}^{r} G_{\sigma}\left(rx, ry\right)F\left(x + rx, y + ry\right)Geo\left(x + rx, y + ry\right)\\
  \label{eq:FinalKernelK}
  \text{with} \enspace k\left(x, y\right) &= \sum_{rx=-r}^{r}\sum_{ry=-r}^{r} G_{\sigma}\left(rx, ry\right)Geo\left(x + rx, y + ry\right),
\end{align}
where $Geo\left(x,y\right)$ represents an indicator function of geometry. It returns 0 if pixel $\left(x, y\right)$ is inside the geometry, 1 otherwise. For the following steps, we decided to use $\sigma=10$ and $r=3\sigma = 30$.

% % % % % % % % % % % % % % % % % % % % % % % % % % % % % % % % %

\subsubsection{Downsampling}

We perform downsampling to reduce the resolution of the image to an acceptable size for the neural network. The transformation of the image is also applied to geometry. There are several ways this step can be performed, such as Fourier or Wavelet transforms. However, we use averaging. Initially, images exported from the CFD solver are at a resolution of 4000x4000 and a single gray scale channel, before being cropped to 3840x3840. The images are then downsampled at the same time as blurred by using the kernel from equation (\ref{eq:FinalKernel}) of size 30 with a stride of 30 without padding. This leads to a processed image size of 128x128. Both input and output of the network operate on this size. Due to the large number of images and high initial resolution, this post-processing was implemented using CUDA allowing to increase throughput by 2-3 orders of magnitude.

% % % % % % % % % % % % % % % % % % % % % % % % % % % % % % % % %
\subsection{Training}\label{sec:TrainingPipeline:Training}
\subsubsection{Hyperparameters}

After identifying the best performing networks (c.f. \autoref{sec:TrainingPipeline:Architecture}), a systematic hyper parameter sweep is performed. The models in the preliminary sweep used the parameters seen in appendix \ref{app:hyperparameter1}. We note from these initial results that networks with larger kernels and a larger number of convolutional channels performed better. This is unsurprising as larger kernels could capture more complex or larger features. Other than network and kernel size, the following parameters seem to be the most promising:

\begin{enumerate}
    \item Beta parameters for the Adam optimizer
    \item Scalar or tensor skip-connection
    \item Skip-connection constraints
    \item Skip-connection location
\end{enumerate}
These are explained in greater detail in the following sections.

\textbf{Beta Parameters}

The Adam optimizer adds an effect akin to "momentum" when updating the weights of the neural network, more specifically, it includes the unbiased first and second moment estimates of previous weight updates. This creates some form of "inertia"---if many past updates tend to move in a specific direction, the current and future updates will be biased to also move in a similar direction. The Adam optimizer is seen having similar benefits to both the AdaGrad and RMSProp extensions of stochastic gradient descent: it is able to deal with sparse data, due to the bias correction when calculating the moving averages, and with noisy, non-stationary data, respectively \cite{Kingma2015AdamAM}.

The two parameters tuned in this phase, $\beta_1$ and $\beta_2$, control the decay rate of these moving averages. We find that the performance of the networks is highly sensitive to changes in both $\beta_1$ and $\beta_2$ and that minor changes to either could result in wildly varying performance when the locations of skip connections were modified. Therefore, we are unable to determine very general rules for better $\beta$ values. However, one general trend is that, with skip connections of depths 1-3, a lower $\beta_1$ value produces good results.

\textbf{Scalar and Tensor Skip-Connections}

In this contribution, skip-connections are implemented in two different ways: scalar (or simple) skip-connections and tensor skip-connections. In reference to figure \ref{fig:unetskip}, the layers $C_{2}$ and $U_{2}$ are both tensors of shape $n \times n \times c$, where $n$ is the length of the down sampled input and $c$ is the number of convolutional channels used in that particular layer. The scalar skip connection combines data using a scalar $t \in [0,1]$ as follows:
\begin{equation*}
    y = tC_{2}+(1-t)U_{2} \quad \text{with} \enspace C_{2},U_{2} \in \mathbb{R}^{n \times n \times c}, \enspace t \in [0,1]
\end{equation*}
The tensor skip connection allows greater degrees of freedom, with a separate scalar value for each channel and the data is combined as follows:
\begin{equation*}
 y = T \circ C_{2}+(1-T) \circ U_{2} \quad \text{with} \enspace C_{2},U_{2} \in \mathbb{R}^{n \times n \times c}, \enspace T \in \mathbb{R}^{c}
\end{equation*}

We experiment with both tensor and scalar skip-connections but there is no clear conclusion on which skip-connection type produced better results. 

\textbf{Constrained Skip-Connections}

Recalling that the values in scalar or tensor skip-connections are constrained between 0 and 1, we can implement this constraint multiple ways. The simplest method is just to clip the value of the skip-connections if they exceed 1 or drop below 0, which we refer to these skip-connections as \textit{constrained}. However, another method of constraining the values is through the use of a sigmoidal function, which maps from the space of real values $\mathbb{R}$ to the closed interval of $[0,1]$. We refer to these skip-connections as \textit{unconstrained}. Here, we found that architectures with unconstrained skip-connections tended to display better results.
 
\textbf{Skip Connection Location}

One more variable we experiment with is the location of skip connections. This means varying the depth at which skip connections are placed. A higher skip connection (at depth 1 or 2) passes larger chunks of information as it connects the earlier layers to the last few layers. Skip connections at lower depths pass smaller amounts of data and bypass a smaller number of convolutional layers.

Our experimentation indicats that placing skip connects at a higher depth typically improves performance. Interestingly, not all skip connections see use during training. There are cases where only two out of three skip connections pass any data, i.e., the parameter $t$ is close to 0. The best performing architecture has skip connections at depths 2, 3, 4, and 5. 

% % % % % % % % % % % % % % % % % % % % % % % % % % % % % % % % %
\subsubsection{Loss Function} \label{sec:nnloss}

\begin{figure}[htp]
	\centering
	\begin{subfigure}[b]{0.40\linewidth}
		\centering
		\includegraphics[width=0.9\linewidth]{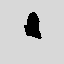}
		\caption{Geometry w/o prism layer}
	\end{subfigure}
		\begin{subfigure}[b]{0.40\linewidth}
		\centering
		\includegraphics[width=0.9\linewidth]{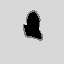}
		\caption{Geometry  w/ prism layer (gray)}
	\end{subfigure}
	\caption{Geometry masks for loss calculation}
	\label{fig:geomask}
\end{figure}

A custom loss function is used in the the training of our neural network. To ensure not to learn inside the geometries, we mask the loss function with the geometry. For example, we compute the $L2$ loss as follows.
\begin{equation}
\label{eq:LossKernel}
  L\left(x, y, Geo\right) = \sum_{i \in len(x)} \left(x_i - y_i \right)^2 \left(1 - Geo_i\right),
\end{equation}
where $Geo$ is an indicator function of the geometry and prism layer for prediction $x$ and ground truth $y$. As we cannot iteratively refine the prism-layers using the Simcenter STAR-CCM+ software, we also mask the prism layers, as seen in figure  \ref{fig:geomask}, to prevent the model from learning the extremely fine mesh density along the prism layers.

%%%%%%%%%%%%%%%%%%%%%%%%%%%%%%%%%%%%%%%%%%%%%%%%%%%%%%%%%%%%%%%%%
\section{Results}\label{sec:Results}
\subsection{Big Data}
\begin{figure}[ht]
	\centering
	\includegraphics[width=0.95\linewidth]{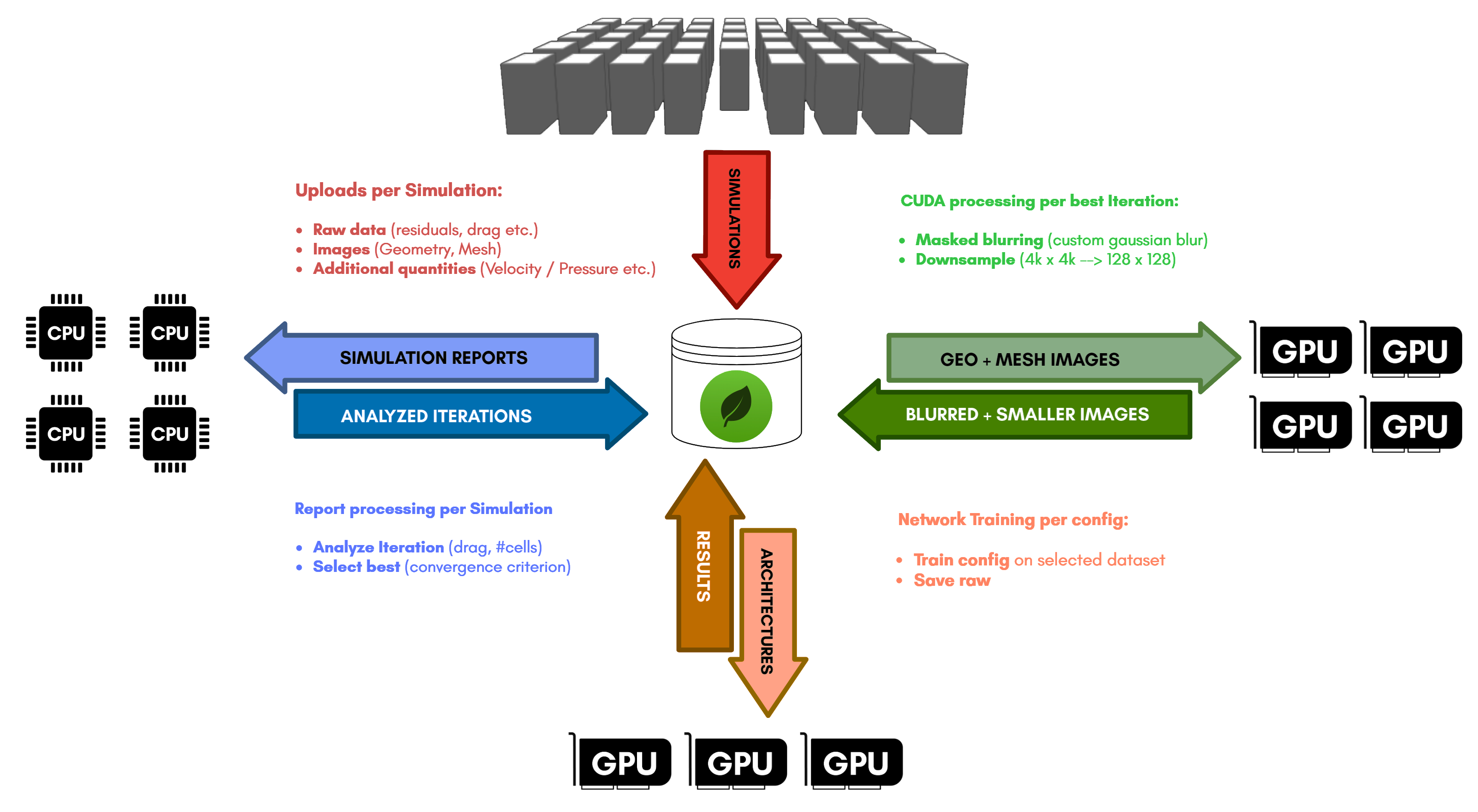}
	\caption{Dataflow from Simulation to trained neural network. Top: LRZ Compute Clusters, Center: Locally hosted MongoDB, Rest: Local Compute Units.}
	\label{fig:bigdataresults}
\end{figure}

Training our neural networks largely relied on the integrity and quality of the data produced in our simulation pipeline. To achieve the amounts of data needed, the process of data acquisition, pre-processing and training had to be mostly automated. Figure \ref{fig:bigdataresults} shows the overall flow of raw data produced on LRZ compute clusters (\url{https://www.lrz.de/english/}), moving through different stages of pre-processing to finally being used as training input for the network architectures.

The pipeline design has 4 stages divided by different compute needs. Raw simulation data is produced with heavy use of parallelism on CPU clusters, with individual runs taking between 30-60 minutes with 10 cores (Figure \ref{fig:bigdataresults} red). Exported data is uploaded to the central element of the pipeline, our locally hosted MongoDB instance handling all data operations from all stages. Generalized pre-processing (Figure \ref{fig:bigdataresults} blue) including assessment of mesh quality is done on local CPU compute units, selecting which refinement iterations to use further down the pipeline. Image pre-processing (Figure \ref{fig:bigdataresults} green) is handled by local GPU compute units using CUDA-accelerated code for selected iterations and also stored back in the database. Finally, all data necessary for training is combined in a single file. This file is then distributed to machines equipped with suitable GPUs training different neural network setups, which in return upload the results in form of the final loss and accuracy of the tested architecture (Figure \ref{fig:bigdataresults} orange).

In total 65.000 simulations were performed on different clusters. At peak production rate, 50 concurrent simulations saturated the IvyMUC\footnote{\url{https://doku.lrz.de/display/PUBLIC/IvyMUC}} cluster at LRZ containing 496 Ivy Bridge CPU. With an average of 4.3 refinement iterations/simulation, 280.000 different meshes were created and their quality assessed. For each mesh, multiple qualities were exported for potential future use, which led to a database size of 1.2 TB containing 3.6 Mio. exported fields. Full simulation files allowing for simulation restarts were not permanently saved, as they amounted to another 5-6 TB. A full dataset containing processed geometries, masks, SDF-fields and mesh targets at 128x128 resolution amounts to 3.3GB in compressed state.

%%%%%%%%%%%%%%%%%%%%%%%%%%%%%%%%%%%%%%%%%%%%%%%%%%%%%%%%%%%%%%%%%%%%%%%%%%%%%%%%%%%%%%%%%%%%%%

\subsection{Neural Network Performance}
\label{sec:nnperformance}

\begin{figure}[h]
	\centering
	\includegraphics[width=0.8\linewidth]{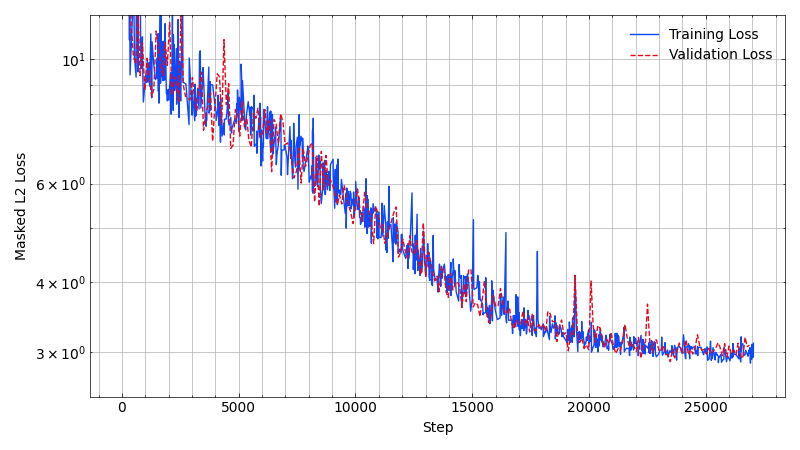}
	\caption{Loss during training of best-performing network excluding geometry and prism layer regions. No overfitting is visible. Both losses have reached a plateau.}
	\label{fig:nntrainingloss}
\end{figure}

For final training, a reduced set containing simulations that had been refined at least 5 or more times were used. This set holds the binary geometry input to the network, a mask to remove geometry and prism layer from the loss function as described in \autoref{sec:nnloss} and the target mesh distribution. All 3 were used at 128x128 resolution. The reduced set amounts to 11.059 samples, split 90\% / 10\% for training and validation. Training was run with batch size 128 and an initial learning rate \num{1e-4} with an exponential decay of 0.89 every 650 steps.  Figure \ref{fig:nntrainingloss} shows the training and validation loss for the final network training run. Both show a continuous decrease until reaching a plateau after 25.000 training steps, which is equivalent to 300 epochs.

\begin{figure}[h]
	\centering
	\includegraphics[width=0.95\linewidth]{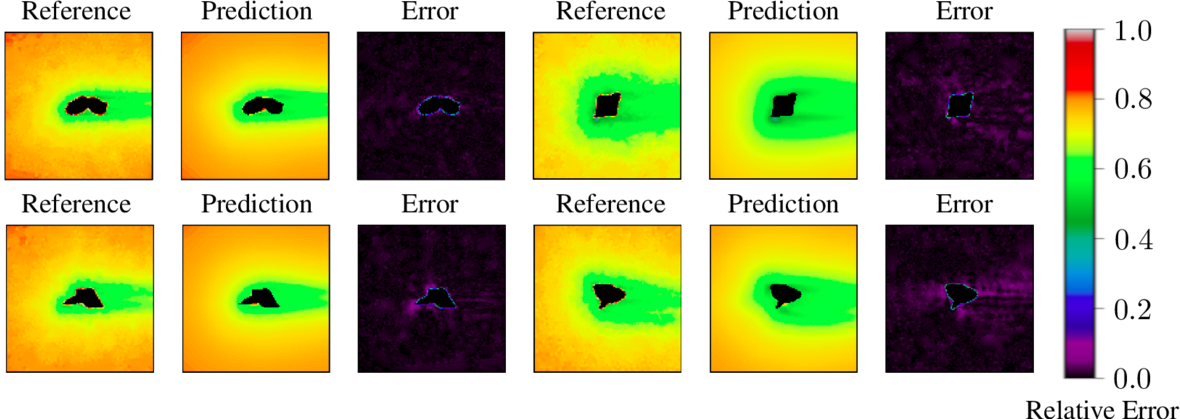}
	\caption{Predictions from neural network, visualized are velocity magnitudes and relative errors between the reference and prediction for different validation data sets.}
	\label{fig:nnresults}
\end{figure}

The best performing neural network is a Staircase UNet with Skip Connections using tensor skip connections. 
This network had a depth of 8 and 2 convolutional layers at each depth. We applied inner tensor skip connections at depths 2,3,4 and 5. We used beta values of $\beta_1=0.9$ and $\beta_2=0.999$ with the exponentially decaying learning rate using an Adam optimizer. This network had $8.5\times10^{7}$ degrees of freedom and obtained an accuracy of $98.7\%$ on the validation dataset. In Figure \ref{fig:nnresults} we illustrate four example geometries with their given reference, the resulting prediction for the velocity magnitude and the corresponding relative error.

%%%%%%%%%%%%%%%%%%%%%%%%%%%%%%%%%%%%%%%%%%%%%%%%%%%%%%%%%%%%%%%%%%%%%%%%%%%%%%%%%%%%%%%%%%%%%%

\subsection{Sample Prediction Validation}
\label{sec:predictionvalidation}

Section \ref{sec:nnperformance} shows the ability of the network to learn refinement maps around a random geometry on a 128x128 scale. For future industrial use, the final step would be to automate a meshing process using these refinement maps to produce a mesh directly e.g. in Simcenter STAR-CCM+. Here, we present an example of how this process might work as well as quantitative performance of the resulting mesh.

\begin{figure}[ht]
	\centering
	\begin{subfigure}[t]{0.40\linewidth}
		\centering
		\includegraphics[width=0.9\linewidth]{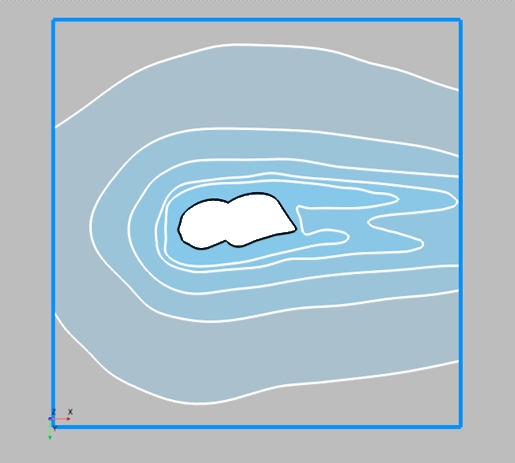}	\caption{Extracted isosurfaces: \num{5e-2}\si{m^2}, \num{1e-2}\si{m^2}, \num{3e-3}\si{m^2}, \num{7.5e-4}\si{m^2}, \num{3e-4}\si{m^2}.}
	\end{subfigure}
	\hspace{0.05\linewidth}
	\begin{subfigure}[t]{0.40\linewidth}
		\centering
		\includegraphics[width=0.9\linewidth]{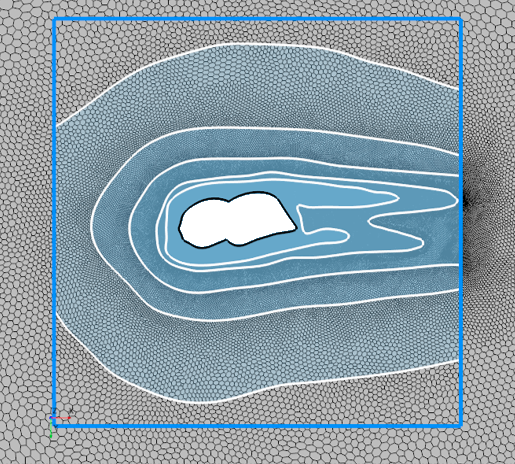}
		\caption{Resulting mesh with volume control zones and growth factor applied.}
	\end{subfigure}
	\caption{Extracted isosurfaces from mesh creation using sample geometry refinement prediction with neural network viewport outlined in blue.}
	\label{fig:nnmesh}
\end{figure}

The neural network output predicts an average mesh cell size per pixel.
Instead of creating the theoretically available 16384 volume controls, we choose to extract 4-5 isosurfaces from the mesh prediction and use those for volume control zones in Simcenter STAR-CCM+ as seen in figure \ref{fig:nnmesh}. 

The network is clearly capturing the general wake behind the geometry as well as the shear layers, which leads to mesh refinement in those critical areas. Given the limitations in capturing the prism layers as mentioned in \autoref{sec:nnloss}, and the limitations imposed by working with downsampled 128x128 data, the network is not predicting the prism layer and smallest cells directly attached to the geometry. This is clearly visible when comparing the adjoint error estimate between the mesh produced using the network prediction and a mesh produced by the iterative refinement discussed in  \autoref{sec:DataPipeline:Adaptivity}.

\begin{figure}
    \centering
    \begin{subfigure}[t]{0.40\linewidth}
		\centering
		\includegraphics[width=0.9\linewidth]{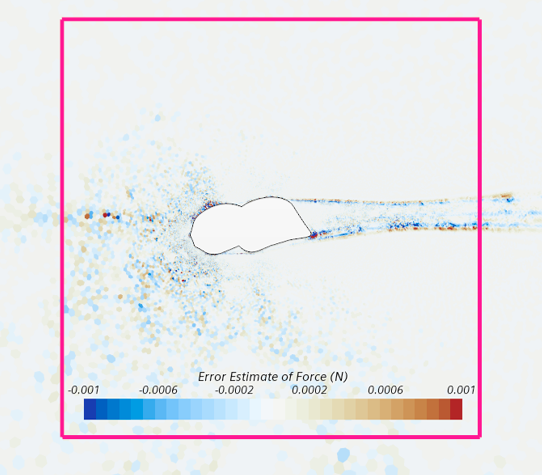}	\caption{Iteratively refined mesh.}
	\end{subfigure}
	\hspace{0.05\linewidth}
	\begin{subfigure}[t]{0.40\linewidth}
		\centering
		\includegraphics[width=0.9\linewidth]{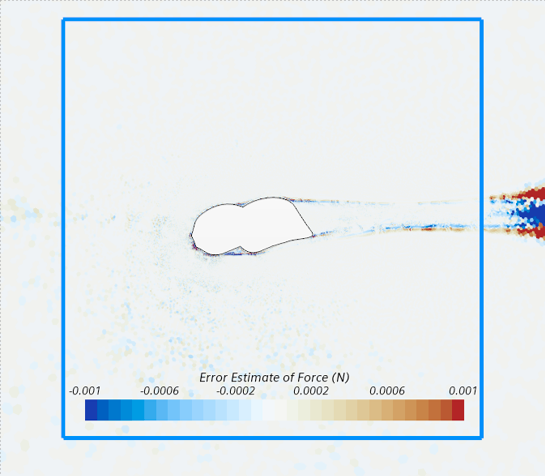}
		\caption{Mesh generated from NN prediction.}
	\end{subfigure}
	\par
	\begin{subfigure}{1.0\linewidth}
	    \centering
    	\begin{tabular}{l|c|c|}
            \cline{2-3}
                                                                  & a) Macro refinement & b) NN refinement \\ \hline
            \multicolumn{1}{|l|}{Drag force (N)}                  & 75.2             & 75.4          \\ \hline
            \multicolumn{1}{|l|}{Sum estimated error in viewport} & -1.4             & -1.9          \\ \hline
            \multicolumn{1}{|l|}{Sum estimated error in domain}   & -1.4             & -1.9          \\ \hline
            \multicolumn{1}{|l|}{Max absolute error in viewport}  & 0.13             & 0.094         \\ \hline
        \end{tabular}
        \caption{Absolute and error value comparison between meshes.}
        \label{tab:nnMeshErrorComparison}
    \end{subfigure}
	\caption{Adjoint error estimate for drag with the neural network viewport shown by square.}
	\label{fig:nnMeshErrorComparison}
\end{figure}

\begin{figure}
    \centering
    \begin{subfigure}[t]{0.40\linewidth}
		\centering
		\includegraphics[width=0.9\linewidth]{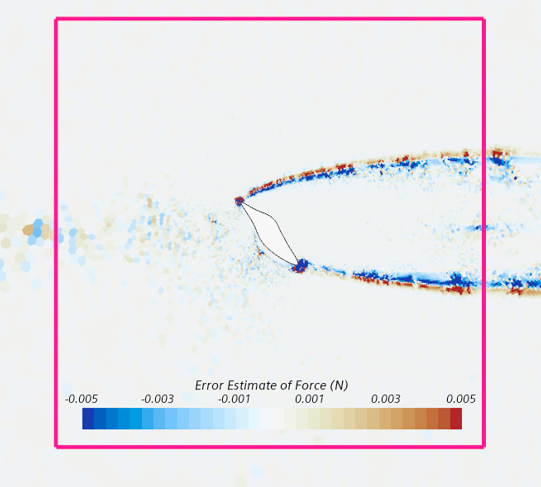}	\caption{Iteratively refined mesh.}
	\end{subfigure}
	\hspace{0.05\linewidth}
	\begin{subfigure}[t]{0.40\linewidth}
		\centering
		\includegraphics[width=0.9\linewidth]{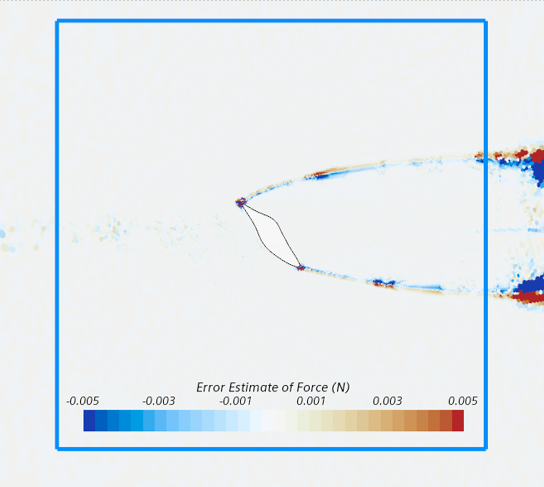}
		\caption{Mesh generated from NN prediction.}
	\end{subfigure}
	\par
	\begin{subfigure}{1.0\linewidth}
	    \centering
        	\begin{tabular}{l|c|c|}
                \cline{2-3}
                                                                      & a) Macro refinement & b) NN refinement \\ \hline
                \multicolumn{1}{|l|}{Drag force (N)}                  & 426.9            & 445.6        \\ \hline
                \multicolumn{1}{|l|}{Sum estimated error in viewport} & -13.3            & -5.0         \\ \hline
                \multicolumn{1}{|l|}{Sum estimated error in domain}   & -15.0            & -8.0         \\ \hline
                \multicolumn{1}{|l|}{Max absolute error in viewport}  & 0.52             & 3.2          \\ \hline
            \end{tabular}
            \caption{Absolute and error value comparison between meshes.}
            \label{tab:nnMeshErrorComparison2}
    \end{subfigure}
	\caption{Adjoint error estimate for drag with the neural network viewport shown by square.}
	\label{fig:nnMeshErrorComparison2}
\end{figure}

Figures \ref{fig:nnMeshErrorComparison} and \ref{fig:nnMeshErrorComparison2} compare the adjoint error estimate on drag force per cell between a mesh iteratively refined by our macro pipeline used for training data production on the left, and the mesh generated from the neural network prediction on the right. Figure \ref{fig:nnMeshErrorComparison}b shows the lack of refinement close to the geometry stemming from the masking of prism layers from our dataset. Compared to the iterative mesh in figure \ref{fig:nnMeshErrorComparison}a, the error especially in front of the geometry is reduced and the wake and shear layers captured. The large error outside the viewport is a result of the default growth rate behaviour of the mesher, as no further control volumes were set outside the neural network viewport. Table \ref{tab:nnMeshErrorComparison} shows similar values for overall drag, as well as overall error in the viewport and total domain with the maximum cell error actually lower for the neural network mesh.

\todo{FM: Can you also add something about the computational cost of the meshes, e.g., {\#} of cells or DoF? Similar for the other table}

Figure \ref{fig:nnMeshErrorComparison2} shows a similar behaviour for a different geometry, which has not been seen during training or validation. Again the error in front of the object is reduced and the shear layers better resolved. The adjoint error estimate in table \ref{tab:nnMeshErrorComparison2} in this case suggests a significantly reduced error in the viewport and overall domain for the neural network mesh. A common issue between the two meshes is the large error in cells around sharp bends around the object. This most likely is a result of the iterative refinement strategy not affecting prism layer settings, so both the iterative mesh and the network prediction based on training data with the same strategy share this weak point.

Overall, the meshes generated using the output of the neural network produce well converging meshes with error estimates comparable to or better than the iterative approach for the two samples investigated. Generally, the network outputs mesh refinement maps with very smooth gradients in cell size, while the iterative approach used to generate the training data often results in meshes with very localized refinement zones. This noisiness in the training data is a major contributor to the limit in accuracy during our network training. Given the network is seemingly not trying to reproduce this noise but rather going for larger structures and smooth gradients along the field bodes well for the usability of predicted meshes.

\todo[inline]{FM: Is there an argument for computational gains? We could e.g. assume the training data production and NN training as offline phase. Then, do we see low computational cost when using the network as a premesher or something similar compared to the usual meshing? If that is true in some way, we should add that as an additional paragraph here (What were the savings? Storage vs. Computations?) and also in the conclusion (Plus there: How can this be improved? Combination of NN with standard meshing?).}

%%%%%%%%%%%%%%%%%%%%%%%%%%%%%%%%%%%%%%%%%%%%%%%%%%%%%%%%%%%%%%%%%
\section{Conclusions and Outlook}\label{sec:Conclusions}
Within this contribution, we presented a novel machine-learning based approach for optimal mesh generation in computational fluid dynamics (CFD) applications. Basis was the commercial computational fluid dynamics simulator Simcenter STAR-CCM+ \cite{StarCCM}, which was expanded by an iterative mesh refinement loop to generate optimized meshes with respect to the overall error. This computational pipeline was based on an appropriate error estimator leveraging adjoint sensitivities\footnote{Meanwhile, corresponding functionality is available in the standard Simcenter STAR-CCM+ solution.}. To achieve optimal results in a machine learning approach, we produced over 60.000 simulations, 6 Terabytes of data, and used about 50 years of serial compute time on CPU.

Based on a subset of this dataset we tested and fine-tuned different neural network architectures relying on three different base architectures all using the same dataset to train on. Automated testing of different variants was performed on multiple GPUs in what would account to roughly 2 months of continuous single GPU training time. Multiple hundred variants were trained and tested and multiple very accurate architectures found. It was shown that the modified U-Net architecture both with and without skip connections is able to predict refined mesh densities for a random geometry with an accuracy of $98$\%. Figure \ref{fig:nnresults} illustrates the predictions obtained by the best performing network.

To produce a neural network of such quality required quite some effort with regard to data creation, processing and training, only made possible by the full automation of the entire process from CFD to Neural Network. The trained neural network that came out of this specific pipeline could be used to give inexperienced CFD users a reasonable first guess for a mesh refinement given their geometry. This could reduce the need for tedious hand-tuning of the mesh as it should predict a mesh that is able to produce a convergent result. This use of neural networks in the general CFD pipeline is one of the most risk-averse ways of using machine learning in this context. Other research focuses on replacing part of the actual CFD solver with neural networks to speed up the simulation. However, this could lead to inaccurate results when dealing with edge cases as the networks itself have no knowledge of the underlying physics. Our approach can at worst produce a bad mesh, which is easily spotted by lack of convergence of the simulation. But it can in the best case significantly speed up the initial phases of setting up the CFD simulation for an inexperienced engineer and actually produce usable results directly after the first run. The compute resources needed for a single prediction would be a matter of seconds on consumer machines and are definitely worth the wait.

So far we have restricted ourselves to very specific setup, i.e. for a fixed speed flow and geometries of a certain size range. Future studies will address the potential to generalize the concept, e.g. predictions for geometry sizes in different orders of magnitude or different flows. Also of high interest would be to scale this approach from 2D to 3D. Though this might lead to an very big amount of data required for training. Leveraging machine learning in the overall CFD workflow of an engineer can speed up tasks that are highly dependent on personal experience like mesh creation. Even more importantly, machine learning in the context of mesh generation has only little influences on validation and verification of simulation methods as required in typical industrial use case.

%%%%%%%%%%%%%%%%%%%%%%%%%%%%%%%%%%%%%%%%%%%%%%%%%%%%%%%%%%%%%%%%%

\noindent\textbf{Acknowledgements:}
The authors greatly acknowledge the contributions of 2018/19 Bavarian Graduate School of Computational Engineering 2018/19 team\footnote{\url{https://www.bgce.de/curriculum/projects/honours-project-2018-2019-machine-learning-and-simulation-for-grid-optimization}} (Peer Breier, Qunsheng Huang, Moritz Krügener, Oleksander Voloshyn, Mengjie Zhao), the support of the Bavarian Graduate School of Computational Engineering\footnote{\url{https://www.bgce.de}}, the Leibnitz Supercomputing Centre providing compute resources, NVIDIA for providing GPUs through their GPU grant program, as well as Siemens AG and Siemens Digital Industry Software for providing extensive support throughout the project.

%%%%%%%%%%%%%%%%%%%%%%%%%%%%%%%%%%%%%%%%%%%%%%%%%%%%%%%%%%%%%%%%%

\bibliographystyle{IEEEtran}
\bibliography{reference}

\begin{thebibliography}{10}
\providecommand{\url}[1]{#1}
\csname url@samestyle\endcsname
\providecommand{\newblock}{\relax}
\providecommand{\bibinfo}[2]{#2}
\providecommand{\BIBentrySTDinterwordspacing}{\spaceskip=0pt\relax}
\providecommand{\BIBentryALTinterwordstretchfactor}{4}
\providecommand{\BIBentryALTinterwordspacing}{\spaceskip=\fontdimen2\font plus
\BIBentryALTinterwordstretchfactor\fontdimen3\font minus
  \fontdimen4\font\relax}
\providecommand{\BIBforeignlanguage}[2]{{%
\expandafter\ifx\csname l@#1\endcsname\relax
\typeout{** WARNING: IEEEtran.bst: No hyphenation pattern has been}%
\typeout{** loaded for the language `#1'. Using the pattern for}%
\typeout{** the default language instead.}%
\else
\language=\csname l@#1\endcsname
\fi
#2}}
\providecommand{\BIBdecl}{\relax}
\BIBdecl

\bibitem{plewa2005adaptive}
T.~Plewa, T.~Linde, V.~G. Weirs \emph{et~al.}, \emph{Adaptive mesh
  refinement-theory and applications}.\hskip 1em plus 0.5em minus 0.4em\relax
  Springer, 2005.

\bibitem{becker2001optimal}
R.~Becker and R.~Rannacher, ``An optimal control approach to a posteriori error
  estimation in finite element methods,'' \emph{Acta numerica}, vol.~10, pp.
  1--102, 2001.

\bibitem{Guo2016}
X.~Guo, W.~Li, and F.~Iorio, ``Convolutional neural networks for steady flow
  approximation,'' in \emph{Proceedings of the 22nd ACM SIGKDD international
  conference on knowledge discovery and data mining}, 2016, pp. 481--490.

\bibitem{Ling2016}
J.~Ling, A.~Kurzawski, and J.~Templeton, ``Reynolds averaged turbulence
  modelling using deep neural networks with embedded invariance,''
  \emph{Journal of Fluid Mechanics}, vol. 807, pp. 155--166, 2016.

\bibitem{Kutz2017}
J.~N. Kutz, ``Deep learning in fluid dynamics,'' \emph{Journal of Fluid
  Mechanics}, vol. 814, pp. 1--4, 2017.

\bibitem{StarCCM}
\BIBentryALTinterwordspacing
{Simcenter STAR-CCM+}, Siemens Digital Industry Software. [Online]. Available:
  \url{https://www.plm.automation.siemens.com/global/en/products/simcenter/STAR-CCM.html}
\BIBentrySTDinterwordspacing

\bibitem{babuvska1987feedback}
I.~Babu{\v{s}}ka and A.~Miller, ``A feedback finite element method with a
  posteriori error estimation: Part i. the finite element method and some basic
  properties of the a posteriori error estimator,'' \emph{Computer Methods in
  Applied Mechanics and Engineering}, vol.~61, no.~1, pp. 1--40, 1987.

\bibitem{becker1996weighted}
R.~Becker and R.~Rannacher, \emph{Weighted a posteriori error control in FE
  methods}.\hskip 1em plus 0.5em minus 0.4em\relax IWR, 1996.

\bibitem{babuvvska1978error}
I.~Babuv{\v{s}}ka and W.~C. Rheinboldt, ``Error estimates for adaptive finite
  element computations,'' \emph{SIAM Journal on Numerical Analysis}, vol.~15,
  no.~4, pp. 736--754, 1978.

\bibitem{babuvska1978posteriori}
I.~Babu{\v{s}}ka and W.~C. Rheinboldt, ``A-posteriori error estimates for the
  finite element method,'' \emph{International Journal for Numerical Methods in
  Engineering}, vol.~12, no.~10, pp. 1597--1615, 1978.

\bibitem{eriksson1995introduction}
K.~Eriksson, D.~Estep, P.~Hansbo, and C.~Johnson, ``Introduction to adaptive
  methods for differential equations,'' \emph{Acta numerica}, vol.~4, pp.
  105--158, 1995.

\bibitem{verfurth1994posteriori}
R.~Verf{\"u}rth, ``A posteriori error estimation and adaptive mesh-refinement
  techniques,'' \emph{Journal of Computational and Applied Mathematics},
  vol.~50, no. 1-3, pp. 67--83, 1994.

\bibitem{ainsworth1993unified}
M.~Ainsworth and J.~T. Oden, ``A unified approach to a posteriori error
  estimation using element residual methods,'' \emph{Numerische Mathematik},
  vol.~65, no.~1, pp. 23--50, 1993.

\bibitem{barth2005posteriori}
T.~J. Barth, ``A posteriori error estimation and mesh adaptivity for finite
  volume and finite element methods,'' in \emph{Adaptive Mesh Refinement-Theory
  and Applications}.\hskip 1em plus 0.5em minus 0.4em\relax Springer, 2005, pp.
  183--202.

\bibitem{dealII}
W.~Bangerth, R.~Hartmann, and G.~Kanschat, ``deal. ii—a general-purpose
  object-oriented finite element library,'' \emph{ACM Transactions on
  Mathematical Software (TOMS)}, vol.~33, no.~4, pp. 24--es, 2007.

\bibitem{FEniCS}
A.~Logg, K.-A. Mardal, and G.~Wells, \emph{Automated solution of differential
  equations by the finite element method: The FEniCS book}.\hskip 1em plus
  0.5em minus 0.4em\relax Springer Science \& Business Media, 2012, vol.~84.

\bibitem{IngridCloud}
\BIBentryALTinterwordspacing
{Ingrid Cloud - Smart Wind Simulations}, Ingrid Cloud. [Online]. Available:
  \url{https://www.ingridcloud.com}
\BIBentrySTDinterwordspacing

\bibitem{becker2002optimal}
R.~Becker, V.~Heuveline, and R.~Rannacher, ``An optimal control approach to
  adaptivity in computational fluid mechanics,'' \emph{International journal
  for numerical methods in fluids}, vol.~40, no. 1-2, pp. 105--120, 2002.

\bibitem{braack2004adaptive}
M.~Braack and A.~Ern, ``Adaptive computation of reactive flows with local mesh
  refinement and model adaptation,'' in \emph{Numerical mathematics and
  advanced applications}.\hskip 1em plus 0.5em minus 0.4em\relax Springer,
  2004, pp. 159--168.

\bibitem{gratsch2006goal}
T.~Gr{\"a}tsch and K.-J. Bathe, ``Goal-oriented error estimation in the
  analysis of fluid flows with structural interactions,'' \emph{Computer
  methods in applied mechanics and engineering}, vol. 195, no. 41-43, pp.
  5673--5684, 2006.

\bibitem{thuerey2018}
N.~Thuerey, K.~Weissenow, H.~Mehrotra, N.~Mainali, L.~Prantl, and X.~Hu,
  ``Well, how accurate is it? a study of deep learning methods for
  reynolds-averaged navier-stokes simulations,'' \emph{arXiv preprint
  arXiv:1810.08217}, 2018.

\bibitem{brunton2020machine}
S.~L. Brunton, B.~R. Noack, and P.~Koumoutsakos, ``Machine learning for fluid
  mechanics,'' \emph{Annual Review of Fluid Mechanics}, vol.~52, pp. 477--508,
  2020.

\bibitem{tompson2017accelerating}
J.~Tompson, K.~Schlachter, P.~Sprechmann, and K.~Perlin, ``Accelerating
  eulerian fluid simulation with convolutional networks,'' in \emph{Proceedings
  of the 34th International Conference on Machine Learning-Volume 70}.\hskip
  1em plus 0.5em minus 0.4em\relax JMLR. org, 2017, pp. 3424--3433.

\bibitem{Giles2000}
\BIBentryALTinterwordspacing
M.~B. Giles and N.~A. Pierce, ``An introduction to the adjoint approach to
  design,'' \emph{Flow, Turbulence and Combustion}, vol.~65, no.~3, pp.
  393--415, 2000. [Online]. Available:
  \url{https://doi.org/10.1023/A:1011430410075}
\BIBentrySTDinterwordspacing

\bibitem{Giannakoglou2008}
K.~C. Giannakoglou and D.~I. Papadimitriou, ``Adjoint methods for shape
  optimization,'' in \emph{Optimization and computational fluid
  dynamics}.\hskip 1em plus 0.5em minus 0.4em\relax Springer, 2008, pp.
  79--108.

\bibitem{mao2016}
X.~Mao, C.~Shen, and Y.-B. Yang, ``Image restoration using very deep
  convolutional encoder-decoder networks with symmetric skip connections,'' in
  \emph{Advances in neural information processing systems}, 2016, pp.
  2802--2810.

\bibitem{Kingma2015AdamAM}
D.~P. Kingma and J.~Ba, ``Adam: A method for stochastic optimization,''
  \emph{CoRR}, vol. abs/1412.6980, 2015.

\end{thebibliography}

% Abstract for the TUM report document
% Included by MAIN.TEX
\clearemptydoublepage

\section*{Appendix}
\begin{appendices}

%%%%%%%%%%%%%%%%%%%%%%%%%%%%%%%%%%%%%%%%%%%%%%%%%%%%%%%%%%%%%%%%%
\section{CFD solver parameters}
 \label{app:CFD_solver}
 
% % % % % % % % % % % % % % % % % % % % % % % % % % % % % % % % %
 
\subsection{Mesh Setup}
 \label{app:CFD_solver_mesh}
 
Table \ref{tab:cfd_mesh_global} states the parameters and their value for the base mesh domain while detailed setups of the prism layer and the wake refinements can be found in table \ref{tab:cfd_prism_layer} and table \ref{tab:cfd_wake_refinemnt}, respectively.

\begin{table}[htp!]
\centering
\begin{tabular}{lll}
Default controls     & Values  & {[}unit{]}        \\ \hline \hline
Base size            & 1       & m                  \\
Target surface size  & 100     & Percentage of base \\
Mininum surface size & 10      & Percentage of base \\
Growth rate          & 1.05    &
\end{tabular}
\caption{Global mesh setup overview}
\label{tab:cfd_mesh_global}
\end{table}

\begin{table}[htp!]
\centering
%\label{tab:cfd_prism_layer}
\begin{tabular}{lll}
Obstacle surface control parameters & Values & {[}unit{]}        \\ \hline \hline
Number of prism layers             & 10     &                    \\
Prism layer total thickness        & 100    & Percentage of base \\
Prism layer stretching           & 1.3    &                      \\
Mininum surface size               & 0.5    & Percentage of base
\end{tabular}
\caption{Obstacle prism layer setup overview}
\label{tab:cfd_prism_layer}
\end{table}

\begin{table}[htp!]
\centering
\begin{tabular}{llll}
               & Wake refinement 1 & Wake refinement 2 & {[}unit{]}         \\ \hline \hline
Isotropic size & 20                & 40                & percentage of Base \\
Distance       & 30                & 70                & m                  \\
Growth rate    & 1.2               & 1.2               &
\end{tabular}
\caption{Wake refinement setup overview} \label{tab:cfd_wake_refinemnt}
\end{table}

% % % % % % % % % % % % % % % % % % % % % % % % % % % % % % % % %
\clearpage
\subsection{Primal Solver}
 \label{app:CFD_solver_primal}
 
The parameters for grid sequencing initialization can be found in table \ref{tab:cfd_grid_sequencing}.

\begin{table}[htp!]
\centering
\begin{tabular}{ll}
Grid sequencing parameters       & Values \\ \hline \hline
Maximum grid levels             & 10     \\
Maximum iterations per level    & 50     \\
Convergence tolerance per level & 0.01   \\
CFL number                      & 50
\end{tabular}
\caption{Grid sequencing setup}
\label{tab:cfd_grid_sequencing}
\end{table}

\begin{figure}[htp]
    \centering
    \begin{subfigure}[b]{0.38\textwidth}
        \centering
        \includegraphics[width=\textwidth]{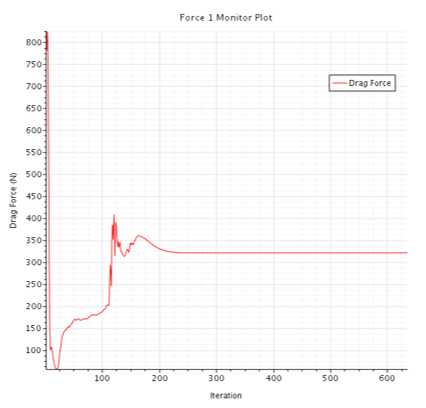}
        \caption[Asymptotic stopping citeria]%
            {{Drag force asymptotic stopping criteria}}
        \label{fig:cfd_stop_cri_asym_drag}
    \end{subfigure}
    \hspace{2mm}
    \begin{subfigure}[b]{0.38\textwidth}
        \centering
        \includegraphics[width=\textwidth]{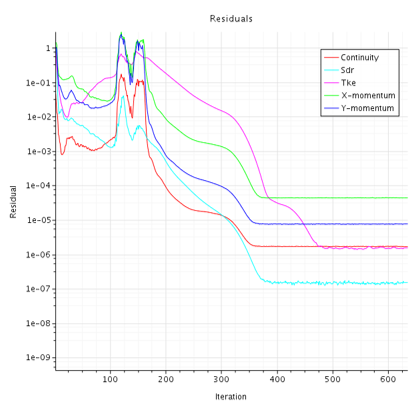}
        \caption[]%
        {{Continuity asymptotic stopping criteria}}
        \label{fig:cfd_stop_cri_asym_cont}
    \end{subfigure}
    \vskip\baselineskip
    \begin{subfigure}[b]{0.38\textwidth}
         \centering
         \includegraphics[width=\textwidth]{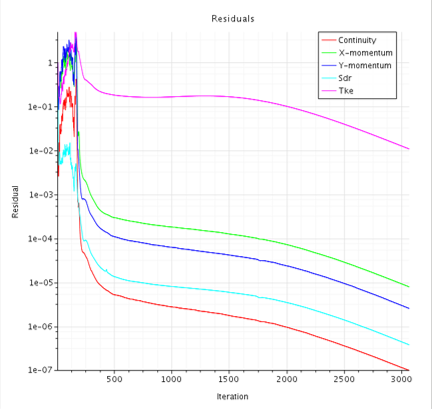}
        \caption[]%
        {{Continuity minimum criteria}}
        \label{fig:cfd_stop_cri_min}
    \end{subfigure}
    \hspace{2mm}
    \begin{subfigure}[b]{0.38\textwidth}
        \centering
        \includegraphics[width=\textwidth]{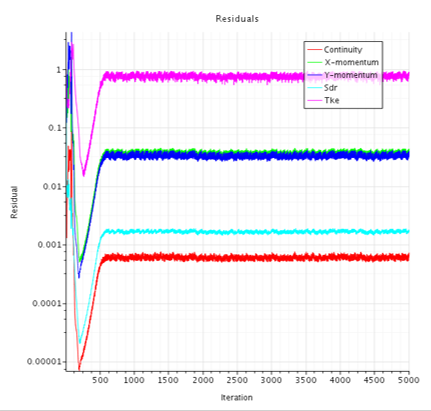}
        \caption[]%
        {{Maximum iteration number criteria}}
        \label{fig:cfd_stop_cri_max}
    \end{subfigure}
    \caption{Visualization of stopping criteria}
    \label{fig:cfd_stopping_criteria}
\end{figure}

To ensure an efficient usage of compute resources as well as easy automation, we consider the following three situations where the simulation should terminate to :

\begin{enumerate}
    \item \textbf{Reaching Steady State:} The ideal case is when the drag force and residuals reach a steady state and the solution does not change with more iterations. During automation, we only consider the continuity residual when checking for steady state as it is representative of the other residual quantities. This case is visualized in figure \ref{fig:cfd_stop_cri_asym_drag} and figure \ref{fig:cfd_stop_cri_asym_cont} where we plot the residuals. We take a moving average of the latest 100 residual values and if this moving average remains within a given tolerance, we accept that the simulation has converged.
    \item \textbf{Reaching Sufficient Accuracy:} Since the adjoint solver only requires a sufficiently converged result, the simulation can also be stopped when the residual falls below a certain value. Similar to the asymptotic stopping criteria, only the continuity residual is considered. We set the threshold for the continuity residual at $10e-7$. One of the example of reaching the minimum criteria can be found in figure \ref{fig:cfd_stop_cri_min} where we can clearly see that the continuity residual (red line) touches the x-axis, which indicates a residual of $10e-7$.
    \item \textbf{Reaching Maximum Iteration Number:} Since a large number of converged simulations is required to form a sufficiently large data set, it is more efficient to drop a simulation that does not converge after a long time and to restart. Therefore, a maximum iteration number criteria is set to avoid wasting resources. The maximum iteration number is set to be 5000 since---after running several batches of random geometries---we observed that stopping criteria 1. and 2. most often took effect before reaching 3000 iterations. Hence, we chose 5000 to include a safety factor to avoid ending viable simulations prematurely. One of the example of reaching the maximum iteration number can be found in figure \ref{fig:cfd_stop_cri_max} where the residuals are not achieving the required threshold.
\end{enumerate}

%%%%%%%%%%%%%%%%%%%%%%%%%%%%%%%%%%%%%%%%%%%%%%%%%%%%%%%%%%%%%%%%%
\clearpage
\section{Explanation Adjoint Method}
  \label{app:adjoint}

A simple explanation of the adjoint method is shown as follows. Given a solution to a series of governing equations $R(F,\omega)$ and a cost function $I(F,\omega)$ both of which vary depending on an arbitrary input parameter $F$ and position $\omega$, we can approximate that gradients of the two variables in relation to changes in F and $\omega$:
\begin{equation}
\label{eq:I}
dI =  \dfrac{\partial I^T}{\partial\omega}d\omega + \dfrac{\partial I^T}{\partial F}dF,
\end{equation}
\begin{equation}
\label{eq:R}
dR = \dfrac{\partial R^T}{\partial\omega}d\omega + \dfrac{\partial R^T}{\partial F}dF = 0.
\end{equation}

Based off equations (\ref{eq:I}) and (\ref{eq:R}), we can formulate the adjoint equation as in equation (\ref{eq:adjoint}), with Lagrange multiplier $\phi$
\begin{equation}
\label{eq:adjoint}
dL = [\dfrac{\partial I^T}{\partial\omega}d\omega- \phi^T\dfrac{\partial R^T}{\partial\omega}d\omega] + [\dfrac{\partial I^T}{\partial F}dF -  \phi^T\dfrac{\partial R^T}{\partial F}dF].
\end{equation}

We then select a Lagrange multiplier such that the first term is zero
\begin{equation}
\label{eq:lagrangemul}
[\dfrac{\partial R}{\partial\omega}]^T\phi = \dfrac{\partial I}{\partial\omega}.
\end{equation}

The resulting equations to solve for the sensitivity of the cost function to changes to parameter $F$ is then
\begin{equation}
\label{eq:partial}
\dfrac{dI}{dF} = [\dfrac{\partial I^T}{\partial F} - \phi^T\dfrac{\partial R}{\partial F}]dF.
\end{equation}

We see that in the equation (\ref{eq:partial}), the sensitivity to parameter $F$ is independent of $d\omega$ and the solution $R$ can be reused when investigating different parameters. Additionally, the equation (\ref{eq:partial}) is linear and is typically far cheaper than solving for $R$.

%%%%%%%%%%%%%%%%%%%%%%%%%%%%%%%%%%%%%%%%%%%%%%%%%%%%%%%%%%%%%%%%%
\clearpage
\section{Hyper-Parameter Tuning: Phase 1}
  \label{app:hyperparameter1}
	\adjustbox{max width=\columnwidth}{
    \begin{tabular}{|c|c|c|c|c|c|c|c|c|c|c|c|c|}
\hline
 &input&mask&reference&depth&channels&kernels&loss&relu&batchnorm&skip type&skip location\\
\hline
 1&sdf&prism&normal&6&64&16x16&L2&.2 down only&middle&none&none\\
\hline
 2&sdf&prism&normal&6&256&4x4&L2&.05 &middle&scalar&all\\
\hline
 3&sdf&prism&normal&7&64&4x4&L2&.2 down only&middle&none&none\\
\hline
 4&sdf&prism&normal&7&64&2: 8x8&L2&.2 &middle&none&none\\
\hline
 5&sdf&prism&normal&7&64&3: 6x6&L2&.2 &middle&scalar&1, 2\\
\hline
 6&sdf&prism&normal&7&128&4x4&L2&.05 down only&middle&tensor&1, 2\\
\hline
 7&sdf&prism&normal&8&64&4x4&L2&.2 down only&middle&none&none\\
\hline
 8&sdf&prism&normal&8&64&8x8&L2&.2 &middle&none&none\\
\hline
 9&sdf&prism&normal&8&64&3: 6x6&L2&.2 &middle&none&none\\
\hline
 10&sdf&prism&normal&8&64&4x4&L2&.05 down only&middle&scalar&1, 2\\
\hline
 11&geo&prism&normal&6&64&16x16&L2&.2 down only&middle&none&none\\
\hline
 12&geo&prism&normal&6&256&4x4&L2&.05 &middle&scalar&all\\
\hline
 13&geo&prism&normal&7&64&4x4&L2&.2 down only&middle&none&none\\
\hline
 14&geo&prism&normal&7&64&2: 8x8&L2&.2 &middle&none&none\\
\hline
 15&geo&prism&normal&7&64&3: 6x6&L2&.2 &middle&scalar&1, 2\\
\hline
 16&geo&prism&normal&7&128&4x4&L2&.05 down only&middle&tensor&1, 2\\
\hline
 17&geo&prism&normal&8&64&4x4&L2&.2 down only&middle&none&none\\
\hline
 18&geo&prism&normal&8&64&8x8&L2&.2 &middle&none&none\\
\hline
 19&geo&prism&normal&8&64&3: 6x6&L2&.2 &middle&none&none\\
\hline
 20&geo&prism&normal&8&64&4x4&L2&.05 down only&middle&scalar&1, 2\\
\hline
 21&sdf&dillute\_prism&normal&6&64&16x16&L2&.2 down only&middle&none&none\\
\hline
 22&sdf&dillute\_prism&normal&6&256&4x4&L2&.05 &middle&scalar&all\\
\hline
 23&sdf&dillute\_prism&normal&6&64&4x4&L2&0&none&tensor&all\\
\hline
 24&sdf&dillute\_prism&normal&6&64&4x4&L1&0&none&tensor&all\\
\hline
 25&sdf&dillute\_prism&normal&7&64&4x4&L2&.2 down only&middle&none&none\\
\hline
 26&sdf&dillute\_prism&normal&7&64&2: 8x8&L2&.2 &middle&none&none\\
\hline
 27&sdf&dillute\_prism&normal&7&64&3: 6x6&L2&.2 &middle&scalar&1, 2\\
\hline
 28&sdf&dillute\_prism&normal&7&128&4x4&L2&.05 down only&middle&tensor&1, 2\\
\hline
 29&sdf&dillute\_prism&normal&7&32&4: 6x6&L2&.2 &middle&scalar&1, 2\\
\hline
 30&sdf&dillute\_prism&normal&7&64&4x4&L2&.2 &middle&scalar& 3, 4, 5, 6\\
\hline
 31&sdf&dillute\_prism&normal&7&64&4x4&L2&0&none&tensor&all\\
\hline
 32&sdf&dillute\_prism&normal&7&64&4x4&L1&0&none&tensor&all\\
\hline
 33&sdf&dillute\_prism&normal&8&64&4x4&L1&0&none&tensor&all\\
\hline
 34&sdf&dillute\_prism&normal&8&64&4x4&L2&.2 down only&middle&none&none\\
\hline
 35&sdf&dillute\_prism&normal&8&32&4: 6x6&L2&.2 &middle&scalar&1, 2\\
\hline
 36&sdf&dillute\_prism&normal&8&64&4x4&L2&.05 down only&middle&none&none\\
\hline
 37&sdf&dillute\_prism&normal&8&64&8x8&L2&.2 &middle&none&none\\
\hline
 38&sdf&dillute\_prism&normal&8&64&3: 6x6&L2&.2 &middle&none&none\\
\hline
 39&sdf&dillute\_prism&normal&8&64&4x4&L2&.05 down only&middle&scalar&1, 2\\
\hline
 40&geo&dillute\_prism&normal&6&64&16x16&L2&.2 down only&middle&none&none\\
\hline
 41&geo&dillute\_prism&normal&6&256&4x4&L2&.05 &middle&scalar&all\\
\hline
 42&geo&dillute\_prism&normal&6&64&4x4&L2&0&none&tensor&all\\
\hline
 43&geo&dillute\_prism&normal&6&64&4x5&L1&0&none&tensor&all\\
\hline
 44&geo&dillute\_prism&normal&7&64&4x4&L2&.2 down only&middle&none&none\\
\hline
 45&geo&dillute\_prism&normal&7&64&2: 8x8&L2&.2 &middle&none&none\\
\hline
 46&geo&dillute\_prism&normal&7&64&3: 6x6&L2&.2 &middle&scalar&1, 2\\
\hline
 47&geo&dillute\_prism&normal&7&128&4x4&L2&.05 down only&middle&tensor&1, 2\\
\hline
 48&geo&dillute\_prism&normal&7&32&4: 6x6&L2&.2 &middle&scalar&1, 2\\
\hline
 49&geo&dillute\_prism&normal&7&64&4x4&L2&.2 &middle&scalar&"3\\
\hline
 50&geo&dillute\_prism&normal&7&64&4x4&L2&0&none&tensor&all\\
\hline
 51&geo&dillute\_prism&normal&7&64&4x4&L1&0&none&tensor&all\\
\hline
 52&geo&dillute\_prism&normal&8&64&4x4&L1&0&none&tensor&all\\
\hline
 53&geo&dillute\_prism&normal&8&64&4x4&L2&.2 down only&middle&none&none\\
\hline
 54&geo&dillute\_prism&normal&8&32&4: 6x6&L2&.2 &middle&scalar&1, 2\\
\hline
 55&geo&dillute\_prism&normal&8&64&4x4&L2&.05 down only&middle&none&none\\
\hline
 56&geo&dillute\_prism&normal&8&64&8x8&L2&.2 &middle&none&none\\
\hline
 57&geo&dillute\_prism&normal&8&64&3: 6x6&L2&.2 &middle&none&none\\
\hline
 58&geo&dillute\_prism&normal&8&64&4x4&L2&.05 down only&middle&scalar&1, 2\\
\hline
 59&sdf&geo&dillute\_prism&7&64&4x4&L2&.2 down only&middle&none&none\\
\hline
 60&sdf&geo&dillute\_prism&7&64&2: 8x8&L2&.2 down only&middle&none&none\\
\hline
 61&sdf&geo&dillute\_prism&7&128&4x4&L2&.05 down only&middle&none&none\\
\hline

 \end{tabular}
 }
 
%%%%%%%%%%%%%%%%%%%%%%%%%%%%%%%%%%%%%%%%%%%%%%%%%%%%%%%%%%%%%%%%%
\clearpage
\section{Hyper-Parameter Tuning: Phase 2}
  \label{app:hyperparameter2}
	\adjustbox{max width=\columnwidth}{
	    \begin{tabular}{|c|c|c|c|c|}
	    \hline
	     &Skip Type & Beta 1& Beta 2& Constrained\\
		 \hline
	     1& scalar & 0.9 & 0.99& yes\\
	     \hline
     	 2& tensor & 0.9 & 0.99& yes\\
     	 \hline
     	 3& scalar & 0.85 & 0.99& yes\\
     	 \hline
     	 4& tensor & 0.85 & 0.99& yes\\
     	 \hline
       	 5& scalar & 0.95 & 0.99& yes\\
       	 \hline
     	 6& tensor & 0.95 & 0.99& yes\\
     	 \hline
	     7& scalar & 0.9 & 0.95& yes\\
	     \hline
     	 8& tensor & 0.9 & 0.95& yes\\
     	 \hline
     	 9& scalar & 0.85 & 0.95& yes\\
     	 \hline
     	 10& tensor & 0.85 & 0.95& yes\\
     	 \hline
       	 11& scalar & 0.95 & 0.95& yes\\
       	 \hline
     	 12& tensor & 0.95 & 0.95& yes\\
     	 \hline
	     13& scalar & 0.9 & 0.999& yes\\
	     \hline
     	 14& tensor & 0.9 & 0.999& yes\\
     	 \hline
     	 15& scalar & 0.85 & 0.999& yes\\
     	 \hline
     	 16& tensor & 0.85 & 0.999& yes\\
     	 \hline
       	 17& scalar & 0.95 & 0.999& yes\\
       	 \hline
     	 18& tensor & 0.95 & 0.999& yes\\     	 
     	 \hline
     	 19& scalar & 0.9 & 0.99& no\\
     	 \hline
     	 20& tensor & 0.9 & 0.99& no\\
     	 \hline
     	 21& scalar & 0.85 & 0.99& no\\
     	 \hline
     	 22& tensor & 0.85 & 0.99& no\\
     	 \hline
       	 23& scalar & 0.95 & 0.99& no\\
       	 \hline
     	 24& tensor & 0.95 & 0.99& no\\
     	 \hline
	     25& scalar & 0.9 & 0.95& no\\
	     \hline
     	 26& tensor & 0.9 & 0.95& no\\
     	 \hline
     	 27& scalar & 0.85 & 0.95& no\\
     	 \hline
     	 28& tensor & 0.85 & 0.95& no\\
     	 \hline
       	 29& scalar & 0.95 & 0.95& no\\
       	 \hline
     	 30& tensor & 0.95 & 0.95& no\\
     	 \hline
	     31& scalar & 0.9 & 0.999& no\\
	     \hline
     	 32& tensor & 0.9 & 0.999& no\\
     	 \hline
     	 33& scalar & 0.85 & 0.999& no\\
     	 \hline
     	 34& tensor & 0.85 & 0.999& no\\
     	 \hline
       	 35& scalar & 0.95 & 0.999& no\\
       	 \hline
     	 36& tensor & 0.95 & 0.999& no\\
     	 \hline
     	 \end{tabular}
	}

\end{appendices}

\end{document}